\documentclass[pdflatex,sn-nature]{sn-jnl}

\usepackage{graphicx}%
\usepackage{multirow}%
\usepackage{amsmath,amssymb,amsfonts}%
\usepackage{amsthm}%
\usepackage{mathrsfs}%
\usepackage[title]{appendix}%
\usepackage{xcolor}%
\usepackage{textcomp}%
\usepackage{manyfoot}%
\usepackage{booktabs}%
\usepackage{algorithm}%
\usepackage{algorithmicx}%
\usepackage{algpseudocode}%
\usepackage{listings}%
\usepackage{booktabs}
\usepackage{tabularx}
\usepackage{array}
\usepackage{enumitem}
\usepackage{caption}

\theoremstyle{thmstyleone}%
\theoremstyle{thmstyletwo}%

\theoremstyle{thmstylethree}%

\newcommand{\modelname}{INOS}

\begin{document}

\title[\modelname{}]{Learning Implicit Solves for Dynamic Simulations}

\author[1]{\fnm{Chawit} \sur{Kaewnuratchadasorn}} \email{chawit.ka@connect.hku.hk}

\author*[1]{\fnm{Jiaji} \sur{Wang}} \email{cewang@hku.hk}

\author[2]{\fnm{Xin} \sur{Nie}} \email{xinnie@tsinghua.edu.cn}

\affil*[1]{\orgdiv{Department of Civil Engineering}, \orgname{The University of Hong Kong}, \orgaddress{\street{Pok Fu Lam}, \city{Hong Kong}, \country{China}}}

\affil[2]{\orgdiv{Department of Civil Engineering}, \orgname{Tsinghua University}, \orgaddress{\city{Haidian}, \state{Beijing}, \country{China}}}

\abstract{High-fidelity simulation of dynamical systems is central to scientific discovery and engineering design, yet remains computationally demanding. Here we introduce the Implicit Neural Operator Solver (\modelname{}), which learns the implicit solve and encodes a time integrator. \modelname{} pretrains a neural operator from static equilibrium solutions and reuses it within Newmark-$\beta$ integration. At each time increment, the integrator forms an effective equilibrium problem, the pretrained neural operator infers the next-step solution field, and the Newmark relations advance the dynamic state. Numerical examples in linear and nonlinear elastodynamics demonstrate accurate full-field predictions without dynamic trajectory supervision. Comparative results show that \modelname{}, evaluated zero-shot, achieves higher predictive accuracy than neural solver baselines trained on dynamic simulations. In addition, \modelname{} delivers over 4,600$\times$ speedup in nonlinear simulations over sequential classical finite-element solvers. \modelname{} natively accommodates changes in dynamic conditions and enables reuse of the pretrained operator across spatial and temporal discretizations. Further evaluation demonstrates a stable million-step rollout. These findings establish \modelname{} as a reusable neural implicit solver for accurate, efficient, and robust dynamic modeling.}


\maketitle

\section{Introduction}
\label{sec:intro}

High-fidelity and efficient simulation of dynamical systems is foundational to advances across mechanics, engineering, and the physical sciences. Classical numerical solvers typically combine spatial discretization methods, including mesh-based \cite{fdm,fem,fvm}, mesh-free and particle-based \cite{sph,md,dem}, and hybrid methods \cite{mpm}, with explicit time integrators \cite{euler,central-diff,rk} or implicit schemes \cite{newmark,hht,gen-alpha}. This versatility, however, is constrained by computational cost at scale, especially for inverse and design-optimization problems.

Scientific machine learning has emerged as a computational paradigm that integrates data-driven approaches with physical knowledge, enabling new capabilities in modeling and understanding physical systems while reducing computational costs. Existing approaches are commonly organized into physics-guided, physics-informed, physics-encoded, and operator-learning paradigms \cite{review-pg-pi-pe,discuss-peml}. Physics-guided models use physical knowledge to guide data-supervised learning through simulation or experimental data, physical features, priors, or model design \cite{pgdl-survey,pgml-sd,pgnn-nl,pgrnn-sdi}. Physics-informed models impose governing equations and initial or boundary conditions through residual-based loss functions \cite{pinn,piml,pidl-sm,pidl-multi,phylstm,phygeonet,xpinn}. Physics-encoded models embed known operators, symmetries, or numerical structures into the architecture \cite{neuralode,lann,phycrnet,percnn}. These methods have enabled progress in forward partial differential equation (PDE) solving, inverse identification, topology and inverse design, sparse-data reconstruction, and constitutive discovery. 

A complementary direction is operator learning, which shifts the learning target from instance-specific predictions to mappings between function spaces, enabling neural operators to approximate physical solution operators \cite{review-pg-pi-pe,ml-pde}. Once trained on a family of input functions, a neural operator can infer full-field solutions without solving the governing equations anew \cite{neuraloperator,neuraloperator-ss}. Foundational neural operator architectures provide different formulations of this principle: Deep Operator Networks (DeepONet) \cite{deeponet} separate the encoding of input functions and query locations through branch–trunk decompositions, and Fourier Neural Operators (FNO) \cite{fno} approximate nonlocal integral kernels in spectral space to enable efficient field-to-field mappings. Subsequent developments have extended operator learning to geometry-aware domains \cite{geofno}, physics-constrained training \cite{pino}, multiple input functions \cite{mionet}, Transformer-based architectures \cite{gnot}, and scalable scientific simulation \cite{gino}. These advances have motivated applications in engineering mechanics, including the discovery of constitutive laws, structural metamodeling, damage detection, nonlinear model updating, and accelerated structural solvers \cite{no-ss,vino,pinos,femu-fno}. Hybrid numerical–operator formulations, including finite element-neural operator (FE-NO) coupling \cite{fem-deeopnet} and neural operator element methods (NOEM) \cite{noem}, further show that learned operators can be embedded within classical discretization frameworks. In these methods, neural operators approximate local solution maps, element-level responses, or decomposed-domain solutions, enabling simulations with higher efficiency.

Dynamic modeling, however, introduces a distinct challenge: long-horizon rollout must preserve stability under variation in loads, time steps, and structural states.  Recent operator-learning studies have increasingly addressed this issue by modifying how neural operators represent temporal evolution. Fourier Neural Operator–Recurrent Neural Network (FNO–RNN) couples field-to-field operator learning with recurrent sequence updates for long-time integration \cite{fno-rnn}. Autoregressive neural operators analyze stability in autoregressive rollout and introduce regularization to reduce error growth \cite{ano}. Recurrent neural operators target teacher-forcing mismatch and compounding long-horizon errors through recurrent training \cite{rno}, while Time-Integrator-embedded DeepONet (TI-DeepONet) shifts the learning target from direct state prediction to instantaneous time-derivative fields advanced by numerical integrators \cite{tideeponet}. The Attention-based Spatio-Temporal Neural Operator (ASNO) adopts an implicit–explicit decomposition inspired by backward differentiation formulas, combining a Transformer encoder that extrapolates the state from its history with a nonlocal attention operator that approximates the implicit solution step \cite{asno}. More closely related to mechanics, time-marching neural operator–finite-element coupling embeds Newmark-type stepping within a PI-DeepONet-FEM domain-decomposition framework, where the neural operator resolves dynamic subdomains rather than the global implicit solve \cite{tmno}. These advances show that temporal structure is essential for dynamic modeling by the operator learning approach. However, these approaches learn temporal evolution or implicit solution steps using dynamic information. This reliance can lead to high training cost, limited generalization across temporal discretizations and rollouts that are sensitive to forcing shifts and out-of-distribution dynamic states, even when physics priors or recurrent structures are embedded \cite{ml-pde,neuraloperator-ss,rno,fno-dyn-challenge}.

Here, we introduce the Implicit Neural Operator Solver (\modelname{}; Fig.~\ref{fig:method}), a physics-encoded neural operator framework for dynamic modeling in which a pretrained operator approximates the global implicit displacement solve. The motivation is to combine the representational capability of operator learning with the structured state updates of implicit time integration. The central observation is that implicit Newmark integration recasts the governing equations, which are second-order in time, as an effective equilibrium problem at each time increment. \modelname{} pretrains a neural operator on static data and reuses it to infer the equilibrium solution at each Newmark step. The dynamic state, time increment, and excitation determine the effective-equilibrium inputs, while the Newmark relations prescribe the kinematic state updates. The main contributions are as follows. (i) A formulation compatible with different neural-operator backbones yields a modular and differentiable dynamic solver in which an operator pretrained on static equilibria approximates the solution during autoregressive rollout (Section~\ref{sec:method}). (ii) The study links the reliability of dynamic rollout to equilibrium prediction accuracy and pretraining coverage, and explains how local inaccuracies can accumulate over successive time increments (Supplementary Notes~\ref{subsec:supp-error-decomp} and \ref{subsec:supp-coverage-condition}). (iii) Zero-shot dynamic evaluations demonstrate full-field accuracy, lower errors than the evaluated trajectory-trained neural baselines, and faster rollout than the FE baseline solvers, including linear and nonlinear elastodynamics using Newton iterations (Sections~\ref{subsec:result-elastodynamic} and~\ref{subsec:analysis-pde}). (iv) \modelname{} naturally accommodates changes in loading histories, initial conditions, and selected time increments. Generalization tests show this capability, with stable rollout over one million simulation steps in the long-horizon test (Section~\ref{subsec:result-generalize}).

\section{Results}
\label{sec:results}
We evaluate \modelname{} on second-order dynamical systems. The benchmarks cover linear and nonlinear elastodynamics across a range of geometries and system sizes. In each benchmark, a neural operator pretrained on static equilibrium data is reused within the Newmark recurrence to predict dynamic responses without supervised training on dynamic trajectories.

\subsection{Learning implicit solves, encoding time integration}
\label{subsec:result-framework}

Elastodynamics describes the motion of deformable solids, with applications in resilient infrastructure \cite{seisgpt}, digital twins \cite{digital-twins-mech-aero}, and soft robotics \cite{soft-robot-embody}. In the reference configuration, the displacement field $\mathbf{u}(\mathbf{X},t)$ satisfies the balance of momentum,
\begin{equation}
    \rho_0\,\partial_{tt}\mathbf{u}
    =
    \nabla_{\mathbf{X}}\!\cdot\mathbf{P}
    + \mathbf{b}_0,
    \label{eq:results-continuum}
\end{equation}
where $\rho_0$ is the reference density, $\mathbf{P}$ is the first Piola-Kirchhoff stress, and $\mathbf{b}_0$ is the body force per unit reference volume. The constitutive law and deformation kinematics determine whether the response is linear or nonlinear.

Spatial discretization and the addition of viscous damping give
\begin{equation}
    \mathbf{M}\ddot{\mathbf{u}}
    + \mathbf{C}\dot{\mathbf{u}}
    + \mathbf{r}_0(\mathbf{u})
    =
    \mathbf{f}_{\mathrm{ext}}(t),
    \label{eq:results-eom}
\end{equation}
where $\mathbf{u}$ now denotes nodal displacements, $\mathbf{M}$ and $\mathbf{C}$ are constant mass and damping matrices, and $\mathbf{r}_0$ is the restoring force. For linear elasticity, $\mathbf{r}_0(\mathbf{u})=\mathbf{K}_0\mathbf{u}$. The construction applies to second-order systems of this form across spatial discretizations with compatible static and dynamic equilibrium representations.

\begin{figure}[t!]
    \centering
    \includegraphics[width=\linewidth]{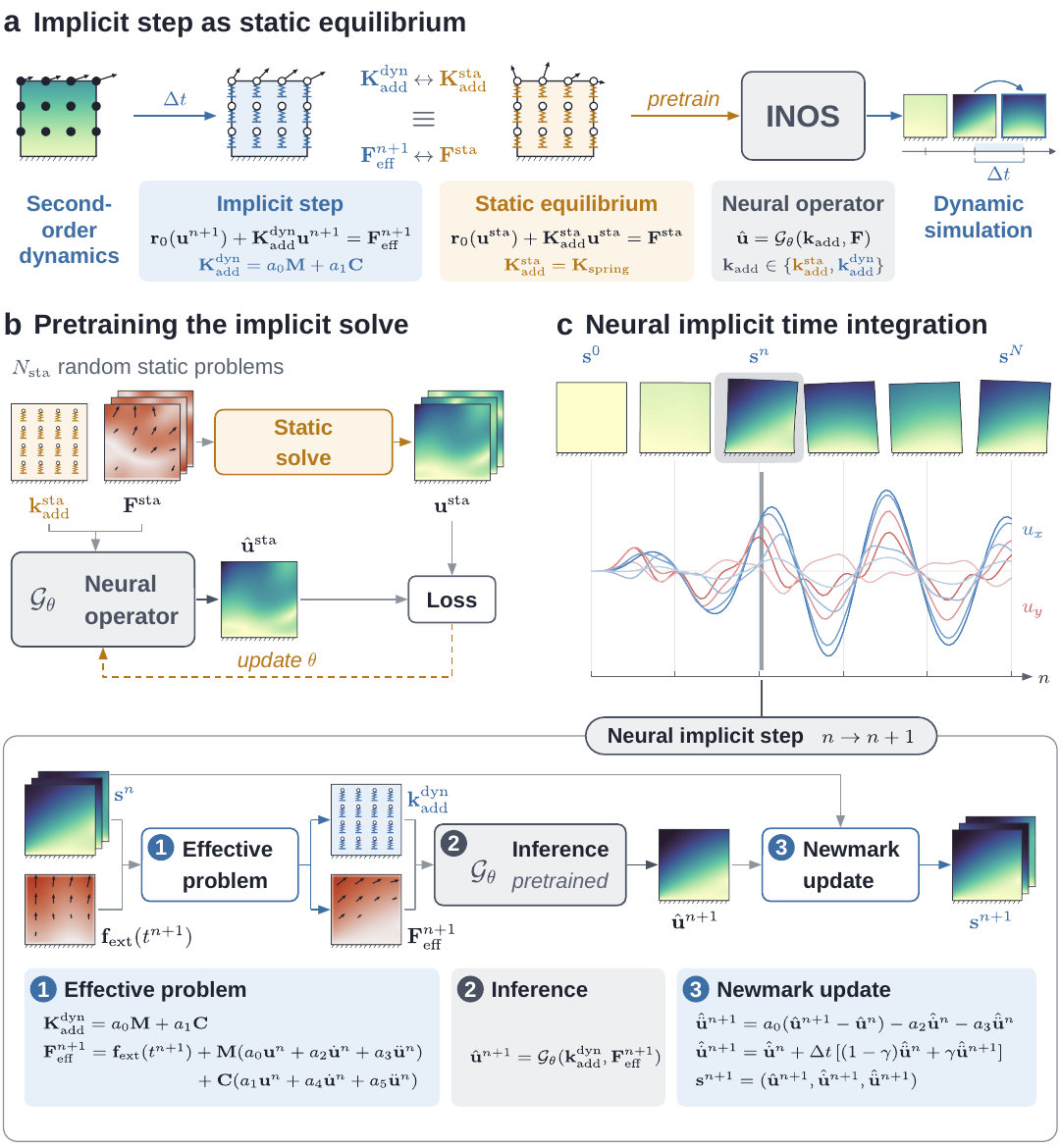}
    \caption{\textbf{Implicit Neural Operator Solver (\modelname{}): learning static equilibrium for zero-shot implicit dynamics.} \textbf{a}, After spatial discretization, Newmark integration reformulates each dynamic step as static equilibrium with added stiffness $\mathbf{K}^{\mathrm{dyn}}_{\mathrm{add}}=a_0\mathbf{M}+a_1\mathbf{C}$, where $\mathbf{M}$ and $\mathbf{C}$ represent mass and damping, and $a_0,a_1$ are Newmark coefficients. \textbf{b}, A neural operator $\mathcal{G}_\theta$ learns the equilibrium map from static problems with added nodal springs, without dynamic training data. \textbf{c}, In autoregressive rollout, the predicted state $\mathbf{s}^{n}$ and external load determine the effective force (1), the operator predicts displacement (2), and the Newmark update advances velocity and acceleration (3). Only the implicit solve is learned through static equilibrium. The time integration follows the prescribed scheme.}
    \label{fig:method}
\end{figure}

Newmark integration yields effective equilibrium problems that share the form of static problems with added nodal springs (Fig.~\ref{fig:method}a):
\begin{equation}
\begin{aligned}
    \mathbf{r}_0(\mathbf{u}^{\mathrm{sta}})
    + \mathbf{K}_{\mathrm{add}}^{\mathrm{sta}}
      \mathbf{u}^{\mathrm{sta}}
    &= \mathbf{F}^{\mathrm{sta}},
    \\
    \mathbf{r}_0(\mathbf{u}^{n+1})
    + \mathbf{K}_{\mathrm{add}}^{\mathrm{dyn}}
      \mathbf{u}^{n+1}
    &= \mathbf{F}_{\mathrm{eff}}^{n+1}.
\end{aligned}
\label{eq:results-equilibrium-transfer}
\end{equation}
Here, $\mathbf{K}_{\mathrm{add}}^{\mathrm{sta}}=\mathbf{K}_{\mathrm{spring}}$ and $\mathbf{K}_{\mathrm{add}}^{\mathrm{dyn}}=a_0\mathbf{M}+a_1\mathbf{C}$, with Newmark coefficients $a_0$ and $a_1$ defined in Methods. Lumped mass and mass-proportional damping make the dynamic added stiffness diagonal, represented by nodal coefficients $\mathbf{k}_{\mathrm{add}}$. For each fixed system, \modelname{} learns the equilibrium map from static data using FNO or Transolver (Fig.~\ref{fig:method}b).

During rollout, Newmark constructs the effective force from the predicted state and the next external load. The operator predicts $\hat{\mathbf{u}}^{n+1}=\mathcal{G}_{\theta}(\mathbf{k}_{\mathrm{add}}^{\mathrm{dyn}},\mathbf{F}_{\mathrm{eff}}^{n+1};\mathbf{X})$, and Newmark updates velocity and acceleration (Fig.~\ref{fig:method}c). Time, step size, and external loading enter through the effective inputs. Only the equilibrium map is learned; input construction and state updates follow the prescribed integrator. Zero-shot rollout therefore requires no training or fine-tuning on dynamic trajectories.

\subsection{Elastodynamics from learned static equilibrium} 
\label{subsec:result-elastodynamic}

\modelname{} enables zero-shot dynamic modeling of linear and nonlinear elastic systems across a range of dimensions, geometries, and sizes. For each system, a neural operator is pretrained on static equilibrium data and reused for inference within the Newmark recurrence without training on dynamic trajectories. We adopted FNO \cite{fno} on structured grids and Transolver \cite{transolver} on irregular meshes. Predictions are compared with finite-element references using the same spatial discretization and Newmark settings. Detailed benchmark configurations and training settings are provided in the Supplementary Information \ref{sec:supp-data-protocols}.

\subsubsection{Linear elastodynamics}
In linear elastodynamics, we evaluate \modelname{} on small-strain isotropic elasticity, described by Hooke's law,
\begin{equation}
    \boldsymbol{\sigma}
    = \mathbb{C}:\boldsymbol{\varepsilon}(\mathbf{u}),
    \qquad
    \boldsymbol{\varepsilon}(\mathbf{u})
    = \frac{1}{2}\left[
        \nabla_{\mathbf{X}}\mathbf{u}
        + \left(\nabla_{\mathbf{X}}\mathbf{u}\right)^{\top}
      \right],
    \label{eq:hooke_law_results}
\end{equation}
where $\boldsymbol{\sigma}$ is the stress tensor, $\boldsymbol{\varepsilon}$ is the infinitesimal strain tensor, and $\mathbb{C}$ is the elasticity tensor. The 2D cases use plane-stress formulations, while the 3D case uses 3D solid elasticity. Mass-proportional damping is added after spatial discretization.

\begin{figure}[t!]
    \centering
    \includegraphics[width=\linewidth]{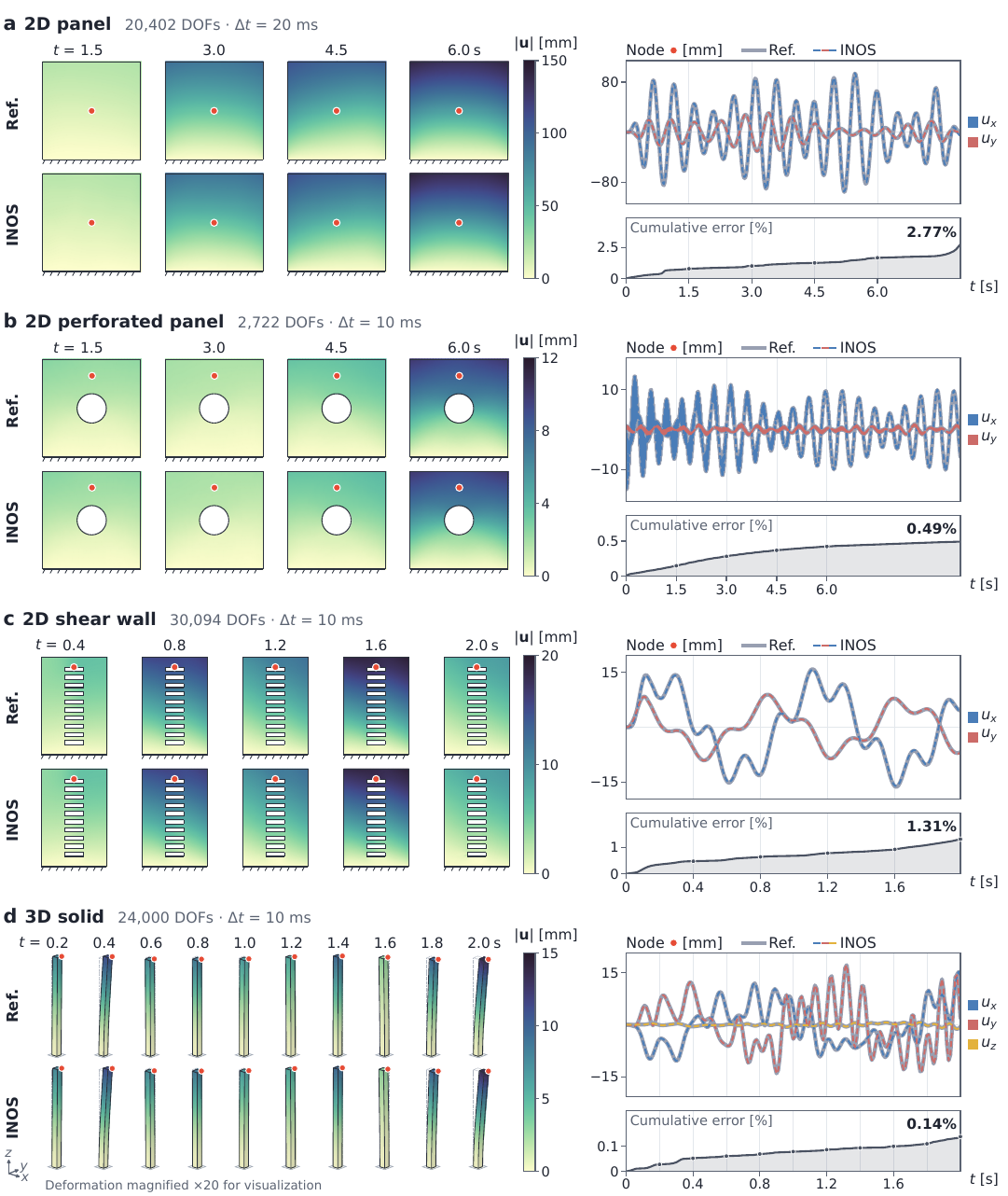}
    \caption{
    \textbf{\modelname{} rollouts on linear elastodynamics.} \textbf{a}, 2D panel. \textbf{b}, 2D perforated panel. \textbf{c}, 2D multistory shear wall. \textbf{d}, 3D solid. Displacement magnitudes $|\mathbf{u}|$, histories at the displayed nodes and cumulative errors are compared with finite-element references (Ref.) using the same spatial discretization, time increment, and Newmark parameters. DOFs, degrees of freedom; $\Delta t$, time increment.}
    \label{fig:linear-elastodynamics}
\end{figure}

For the 2D panel (Fig.~\ref{fig:linear-elastodynamics}a), \modelname{} reproduces the spatiotemporal displacement fields with a relative $\ell_2$ error of $2.77\%$. The predictions capture the spatial displacement patterns and closely follow the FEM reference. For the perforated panel (Fig.~\ref{fig:linear-elastodynamics}b), the learned responses reproduce displacement variations around the internal opening on an irregular mesh, with an error of $0.49\%$. At a larger scale, the multistory shear wall contains $30{,}094$ degrees of freedom (Fig.~\ref{fig:linear-elastodynamics}c). The predicted fields capture displacement variations along the height and around the repeated openings, and nodal histories remain close to the reference, yielding an error of $1.31\%$. We further evaluate \modelname{} on a 3D solid with $24{,}000$ degrees of freedom (Fig.~\ref{fig:linear-elastodynamics}d). The predicted deformation fields and probe histories closely follow the FEM reference, with a full-field relative $\ell_2$ error of $0.14\%$. These results demonstrate the applicability of learned implicit solves or static-to-dynamic transfer to irregular geometries and structural-scale domains.

\subsubsection{Nonlinear elastodynamics} 

\begin{figure}[t!]
    \centering
    \includegraphics[width=\linewidth]{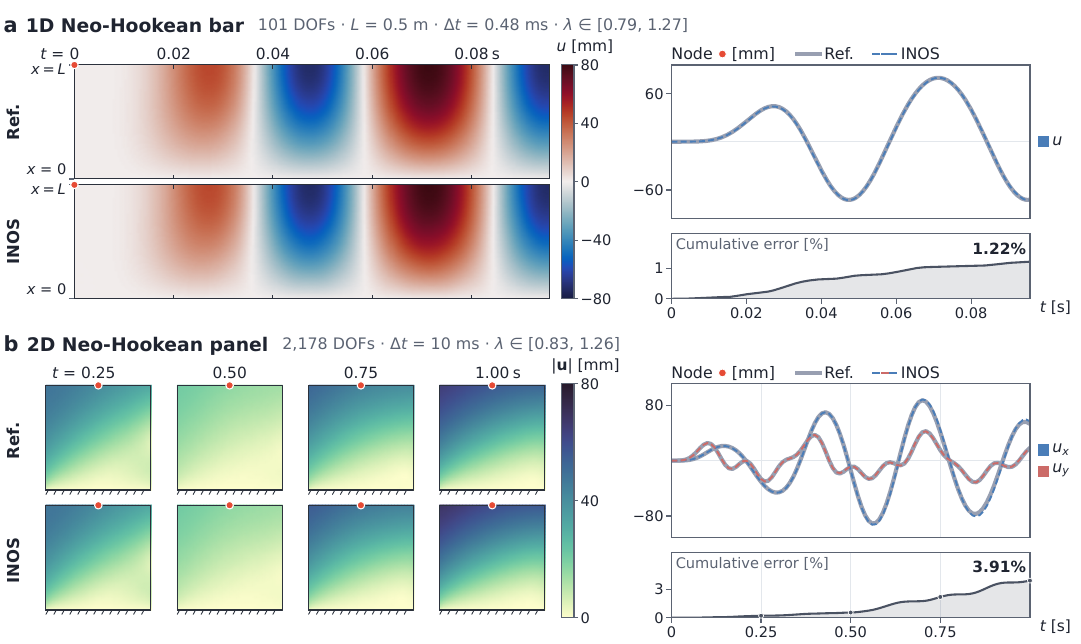}
    \caption{
    \textbf{\modelname{} rollouts on nonlinear elastodynamics.} \textbf{a}, 1D Neo-Hookean bar: axial displacement $u(x,t)$.
    \textbf{b}, 2D Neo-Hookean panel: displacement magnitude $|\mathbf{u}|$ at reported snapshots. Fields, nodal histories, and cumulative errors compare \modelname{} with Newton FE references (Ref.) using the same spatial
    discretization and Newmark parameters.  $\lambda$, reference principal stretches over the rollout; $L$, bar length; $\Delta t$, time increment; DOFs, degrees of freedom.}
    \label{fig:nonlinear-elastodynamics}
\end{figure}

\modelname{} recovers nonlinear dynamics by learning the nonlinear equilibrium map from static equilibrium data. For hyperelastic systems, the restoring force derives from the discretized elastic energy,
\begin{equation}
    \mathbf{r}_0(\mathbf{u})
    = \nabla_{\mathbf{u}}\Psi_{\mathrm{el}}(\mathbf{u}),
    \qquad
    \Psi_{\mathrm{el}}(\mathbf{u})
    = \int_{\Omega_0}
      W\bigl(\mathbf{F}(\mathbf{u})\bigr)\,\mathrm{d}V,
    \label{eq:nonlinear_energy_results}
\end{equation}
where $W$ is the strain-energy density per unit reference volume, $\mathbf{F}$ is the deformation gradient, and $\Omega_0$ is the reference domain. We consider 1D nonlinear axial motion and 2D in-plane motion of a compressible Neo-Hookean solid under plane strain, with constitutive details provided in the Supplementary Information \ref{subsubsec:supp-nonlinear-elastic}. Static labels are generated using Newton's method, and the pretrained operators are reused within the Newmark recurrence through the same input interface. Each nonlinear implicit displacement solve is approximated by a single operator evaluation, without  Newton iterations or training on dynamic trajectories. In one dimension (Fig.~\ref{fig:nonlinear-elastodynamics}a), the predicted displacement field captures the propagation of the nonlinear FEM response. In two dimensions (Fig.~\ref{fig:nonlinear-elastodynamics}b), the predictions reproduce the spatial displacement patterns and closely follow the reference histories in both components, with the larger full-field error occurring in the horizontal component. For each nonlinear benchmark, pretraining and rollout use a fixed added-stiffness field matched to the chosen $\Delta t$. This field remains an operator input.

Across the evaluated systems, \modelname{} demonstrates good performance in zero-shot linear and nonlinear dynamic modeling. Each neural operator learns the equilibrium map, while the prescribed Newmark recurrence advances the dynamic state. We next assess this construction against neural solvers trained with dynamic trajectory data for modeling accuracy and classical implicit solvers for computational cost and efficiency.

\subsection{Analysis of PDE solvers}
\label{subsec:analysis-pde}

This section evaluates \modelname{} against two established approaches to dynamic simulation. Neural solvers trained on dynamic trajectories offer efficient inference, while classical implicit solvers provide the numerical references for assessing accuracy. \modelname{} aims to combine these strengths through static equilibrium learning. Comparisons with neural solvers assess prediction accuracy and robustness to distribution shifts at comparable inference times. Because \modelname{} is trained only on static equilibrium data, every dynamic test is zero-shot with respect to trajectory supervision. Comparisons with classical implicit solvers quantify reductions in simulation cost alongside the approximation errors documented in the previous section.

\subsubsection{Neural solvers}
\label{subsubsec:neural-solvers}

We evaluate \modelname{} against baseline neural solvers on a task to predict displacement responses subjected to external forcing in second-order dynamical systems. The neural baselines comprise FNO~\cite{fno}, Transolver~\cite{transolver}, and recurrent models~\cite{lstm,convlstm}, trained on dynamic trajectories, while \modelname{} is trained on static equilibrium data. For our approach, we also report using FNO and Transolver in both formulations to provide comparisons within the same backbone families. Independent and identically distributed (IID) dynamic trajectories form separate training and test sets: the training set supplies trajectory supervision to the baselines, and the held-out test set is shared by all models. All models are also evaluated on two out-of-distribution (OOD) test sets: force amplitudes beyond the training range, OOD(F), and simulation horizons beyond the training horizon, OOD(H). We assess static-data coverage against the effective-equilibrium inputs associated with the IID training trajectories (Supplementary Notes \ref{subsec:supp-static-data}, \ref{subsec:supp-dynamic-regimes}, and \ref{subsec:supp-data-coverage}). All three test settings are zero-shot for \modelname{} with respect to dynamic trajectory supervision.

\begin{figure}[h!]
    \centering
    \includegraphics[width=\linewidth]{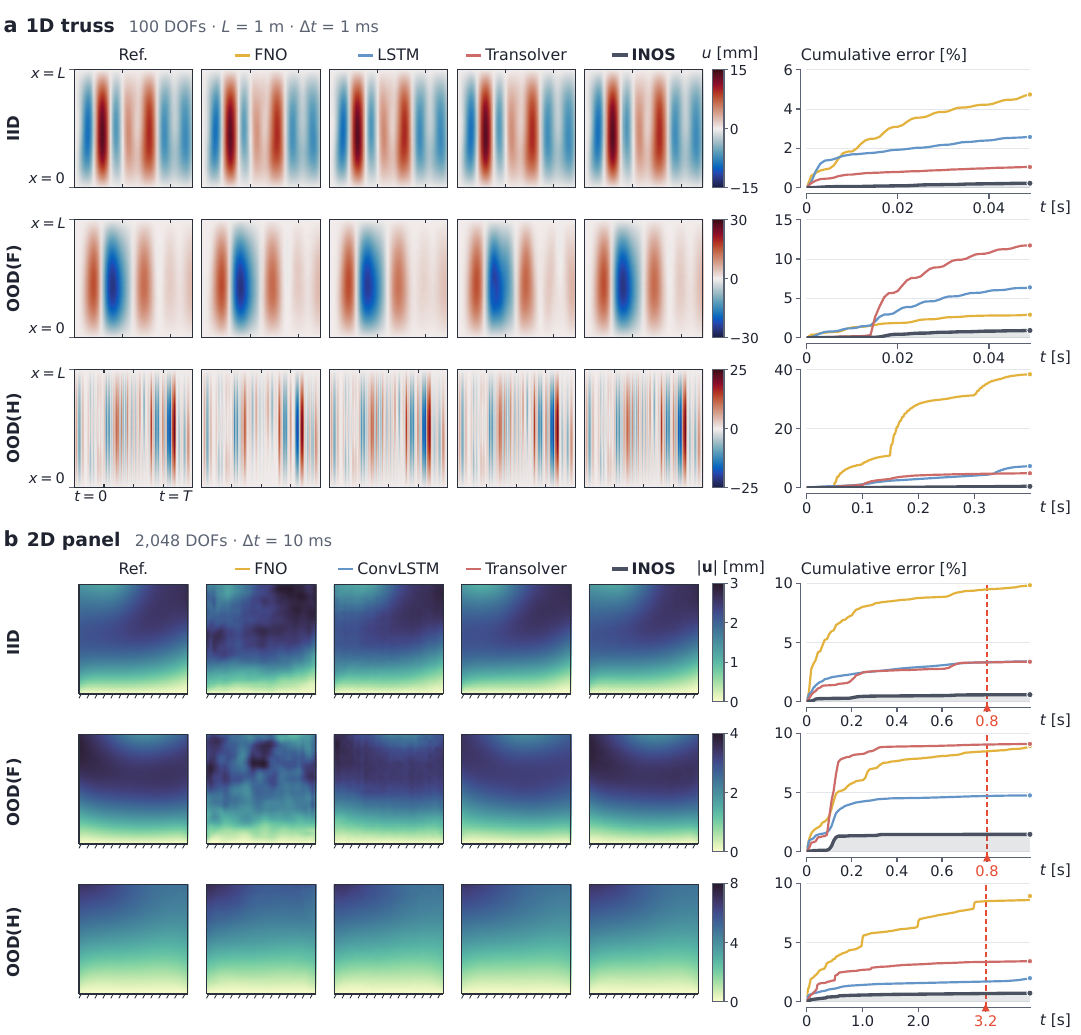}
    \caption{\textbf{Comparison with neural solver baselines trained on dynamic data.} \textbf{a}, 1D dynamics: axial displacement $u(x,t)$. \textbf{b}, 2D dynamics: displacement magnitude $|\mathbf{u}|$; vertical dashed lines mark snapshot times. \modelname{} uses static pretraining and is evaluated zero-shot on dynamic problems, whereas FNO, LSTM/ConvLSTM and Transolver are trained on dynamic trajectories. Tests are IID (independent and identically distributed) or OOD (out of distribution) in force amplitude (F) or rollout horizon (H), relative to the baselines' training distribution. Ref., FE reference; DOFs, degrees of freedom.}
    \label{fig:neural-comparison}
\end{figure}

Fig.~\ref{fig:neural-comparison} shows example displacement predictions and cumulative errors, while Tables~\ref{tab:neural-1d} and~\ref{tab:neural-2d} summarize trajectory-level accuracy. The tested FNO learns a spatio-temporal mapping and shows limited generalization to extended horizons, even with windowed inference. Autoregressive baselines perform better in these horizon tests. The Transolver baseline combines an attention-based neural operator with sequential prediction, without carrying recurrent hidden states, and achieves the lowest horizon-extrapolation errors among the trajectory-trained models. With the Newmark recurrence prescribed, \modelname{} learns only the equilibrium solve and achieves lower errors than its trajectory-trained counterpart within both backbone families across the tested settings. Under force extrapolation, however, the 1D \modelname{} (Transolver) remains less accurate than trajectory-trained FNO. These results demonstrate accurate dynamic prediction from static equilibrium training at inference times comparable to those of trajectory-trained models.

\subsubsection{Classical implicit solvers}
\label{subsubsec:classical-solvers}

Implicit time integration allows larger time steps when explicit schemes are limited by stability, although adequate temporal resolution is still required for accuracy~\cite{newmark}. Each step requires an equilibrium solve, which generally involves repeated iterations for nonlinear systems. \modelname{} approximates this solve using a neural operator trained on static equilibrium data within the prescribed Newmark recurrence. We compare its rollout cost with classical implicit FE solvers for 2D linear and nonlinear elastodynamics under the same spatial discretizations and temporal settings. These comparisons are considered alongside the FEM-reference
errors reported in the previous section, with data generation
and pretraining treated as offline costs.

\begin{figure}[h!]
    \centering
    \includegraphics[width=\linewidth]{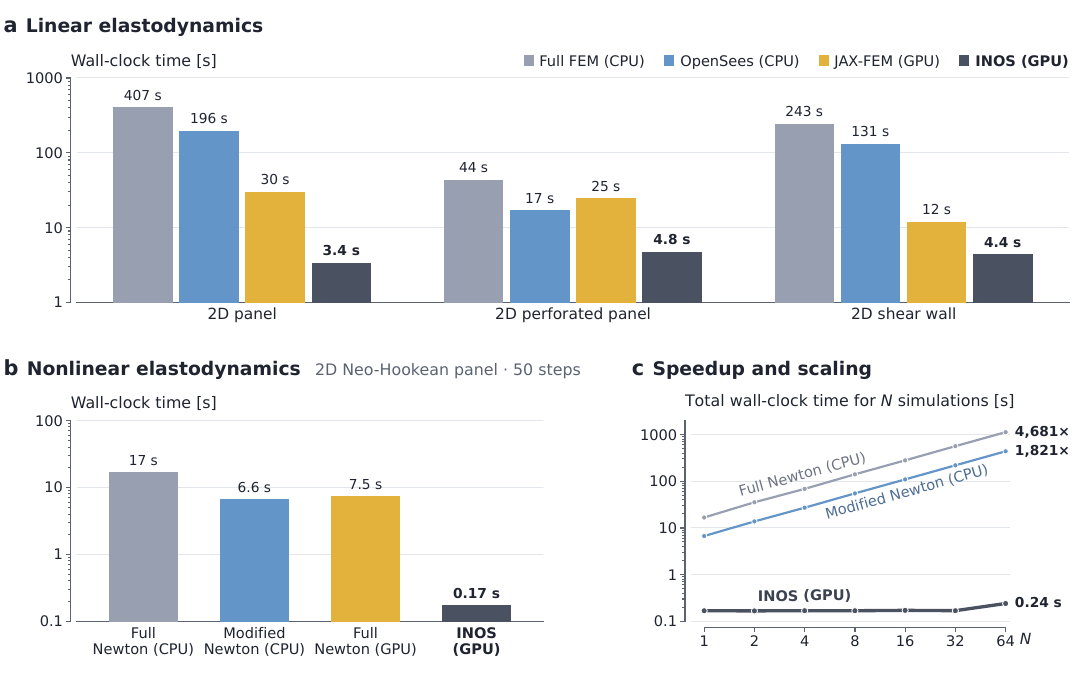}
    \caption{\textbf{Computational efficiency of \modelname{}.} \textbf{a}, Wall-clock rollout times for the 2D linear benchmarks in Fig.~\ref{fig:linear-elastodynamics}. \textbf{b}, Wall-clock times for 50-step rollouts of the 2D Neo-Hookean panel. \textbf{c}, Total wall-clock time for $N$ nonlinear simulations, run sequentially by the FE solvers and as a single GPU batch by \modelname{}. Annotations at $N=64$ show \modelname{} speedups over the FE solvers.}
    \label{fig:computational-efficiency}
\end{figure}

For 2D linear elastodynamics, Fig.~\ref{fig:computational-efficiency}a compares runtimes for dynamic simulations in the panel, perforated domain, and multistory wall. The FEM baseline and OpenSees \cite{opensees} run on the CPU, while JAX-FEM \cite{jax-fem} and \modelname{} run on the GPU. Across these systems, \modelname{} achieves speedups of up to $120\times$ over the FEM baseline. These reductions accompany the close agreement with FEM reference responses shown in Fig.~\ref{fig:linear-elastodynamics}.

For nonlinear elastodynamics, the classical solvers use Newton iterations to satisfy equilibrium at each time step. Full Newton updates the tangent matrix at every iteration, whereas modified Newton reuses the tangent and its factorization across iterations. \modelname{} replaces these iterations with one neural equilibrium evaluation per step. Its GPU rollout achieves speedups of up to approximately $100\times$ over the CPU full Newton implementations and $44\times$ over GPU full Newton (Fig.~\ref{fig:computational-efficiency}b). Batched GPU inference further improves many-query throughput, reaching a speedup of approximately $4{,}700\times$ over sequential CPU full Newton for $64$ simulations (Fig.~\ref{fig:computational-efficiency}c). Together with the response comparisons in Fig.~\ref{fig:nonlinear-elastodynamics}, these results demonstrate efficient nonlinear dynamic simulation through static equilibrium learning, without online Newton iterations.

\subsection{Generalization and rollout stability}
\label{subsec:result-generalize}

\subsubsection{Regimes} 
We evaluate the flexibility of \modelname{} across excitation patterns, initial conditions, time increments, and rollout durations using the 1D linear elastodynamic system. Fig.~\ref{fig:generalization-stability}a shows three response regimes: motion driven by sustained forcing, free vibration initiated by a prescribed displacement, and forced motion followed by load removal. All three simulations use the same equilibrium map learned by the neural operator without retraining. The simulations differ in their initial states and external forces. After load removal, the inertial and damping contributions of the evolving state continue to supply the effective force. At a time increment of $1$~ms, the full-field relative $\ell_2$ errors against the FEM references are below $0.5\%$ in all three cases. These results show that \modelname{} can simulate forced motion, free vibration, and the transition from forced to free motion by combining a neural representation of static equilibrium with encoded Newmark time integration, without dynamic trajectory supervision.

\subsubsection{Time increments}
We evaluate reuse of the pretrained equilibrium operator across time increments, with $\Delta t$ held fixed within each rollout. Changing $\Delta t$ updates the Newmark coefficients used to construct the added stiffness and effective force and to advance the velocity and acceleration, while the pretrained operator remains unchanged. Fig.~\ref{fig:generalization-stability}a compares increments between $0.8$ and $10$~ms in different cases.  Full-field relative $\ell_2$ errors remain below $1\%$ in all tested cases without retraining. These results demonstrate reuse across the tested temporal discretizations within the effective-input range covered by static pretraining (Supplementary Information~\ref{sec:supp-coverage}), without requiring dynamic trajectory data at each time increment.

\subsubsection{Rollout stability}
\begin{figure}[h!]
    \centering
    \includegraphics[width=\linewidth]{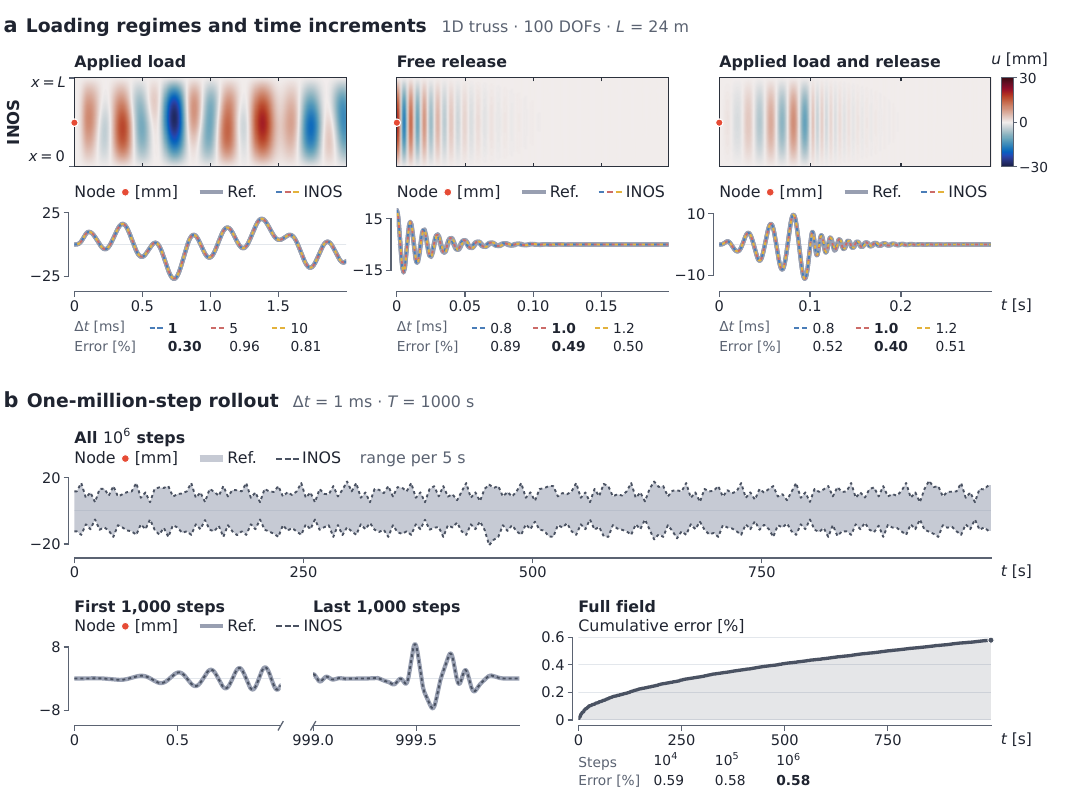}
    \caption{\textbf{Generalization and rollout stability.} \textbf{a}, 1D dynamical responses under applied loading, free release from an initial condition, and loading followed by release, at the indicated time increments $\Delta t$. Space-time maps show \modelname{} predictions at $\Delta t=1$ ms; nodal histories compare predictions with
    FE references (Ref.) at the marked node. \textbf{b}, A $10^6$-step rollout at $\Delta t=1$ ms: nodal displacement ranges within 5-s windows, the first and last 1,000 steps, and cumulative full-field error. All predictions were obtained from the same pretrained \modelname{} solver. DOFs, degrees of freedom.}
    \label{fig:generalization-stability}
\end{figure}

The same pretrained solver is used for an autoregressive rollout of one million time steps, corresponding to $1{,}000$~s of simulated motion (Fig.~\ref{fig:generalization-stability}b). Each predicted state supplies the effective-equilibrium inputs for the next step, allowing approximation errors to propagate through the recurrence. The predicted displacement fields and midpoint response closely match the FEM reference, with a full-field relative $\ell_2$ error of $0.58\%$ over the complete rollout. The cumulative error grows gradually throughout the simulation. These results demonstrate empirical rollout stability over the tested horizon and support accurate long-duration dynamic prediction from static equilibrium training without retraining or dynamic trajectory supervision.

\section{Discussion}
\label{sec:discussion}

This work introduces \modelname{}, a neural implicit solver that learns an equilibrium representation from static data for dynamic modeling of second-order systems. We embed a pretrained neural operator within a prescribed Newmark-$\beta$ time integrator. At each implicit step, the integrator constructs an effective equilibrium problem, and the neural operator predicts the displacement at the next time step. The integrator then updates the dynamic state to continue the autoregressive rollout. From this perspective, \modelname{} is a physics-encoded neural operator that employs learned static equilibrium representations for dynamic prediction through encoded time integration. The added-stiffness and force inputs parameterized during static pretraining allow the learned solver to accommodate variations in the time increment, external excitation, and dynamic state. Once the equilibrium map is learned, temporal evolution follows the prescribed recurrence without fine-tuning on dynamic trajectories. Because the construction operates on the semi-discrete equations of motion, compatible static training data can be obtained from different spatial discretizations.

Across linear and nonlinear elastodynamics, \modelname{} closely reproduced FE reference solutions on structured and unstructured meshes, without dynamic trajectory supervision. This static-to-dynamic transfer demonstrates that equilibrium data can support accurate dynamic modeling when temporal evolution is supplied by a prescribed time integrator. Lower rollout errors than trajectory-trained counterparts within both adopted backbones in the neural operator family, at similar inference times, further support the value of this construction across different operator architectures. The nonlinear results are particularly significant: the learned operator approximates a nonlinear equilibrium map, replacing repeated online Newton iterations with neural evaluations. The observed runtime reductions relative to classical implicit solvers make this approach especially relevant to repeated analyses, where the costs of static data generation and pretraining can be amortized. In \modelname{}, a pretrained neural operator map supports different loading regimes, initial conditions, and time increments, and maintains close agreement with FEM over one million steps. These capabilities reflect the central contribution of \modelname{}: learning the equilibrium solve while retaining prescribed temporal evolution, allowing static training to support accurate, efficient, and flexible dynamic simulation. Although we demonstrated the performance in elastodynamics, \modelname{} could be applied to other second-order dynamical systems whose implicit steps admit a similar equilibrium form, with static pretraining tailored to each system.

\modelname{} can be extended to a broader range of structures, dynamical systems, spatial resolutions, and time increments by adapting its equilibrium formulation and static training family. Static pretraining should represent the effective forces and added stiffnesses expected during rollout. Coverage checks guide static data preparation, while rollout validation assesses cumulative errors (Supplementary Information \ref{sec:supp-coverage}). In this work, each neural operator represents a fixed structural configuration; parameterizing geometry, material properties, and boundary conditions could extend transfer across structures. Higher spatial derivatives can be accommodated through compatible spatial discretizations and restoring-force models. For first-order dynamical systems, compatible implicit schemes recover the same equilibrium form with added terms proportional to $\mathbf{M}/\Delta t$, while higher temporal orders require appropriate implicit formulations and training families. The selected neural operator backbones also allow evaluation on refined meshes without retraining, although accuracy decreases with refinement, motivating investigation of consistent nodal input scaling and training across multiple resolutions (Supplementary Information \ref{sec:supp-resolution-transfer}). The nonlinear experiments use a fixed added-stiffness field; sampling this input could extend nonlinear simulations to different time increments. With lumped mass and mass-proportional damping, updating the effective inputs for a new time increment requires $\mathcal{O}(N)$ work for $N$ degrees of freedom and no matrix refactorization. Coupling this flexibility with error estimates that account for both temporal discretization and neural approximation could support adaptive stepping within a validated input range.

\modelname{} provides a modular route from static analysis to dynamic simulation. Existing static modeling capabilities can thus support ranges of dynamic analyses. This structure could benefit workflows such as seismic risk assessment, uncertainty quantification, and design optimization. Future work could exploit the end-to-end differentiable rollout for load identification and, with relevant parameters represented in the operator inputs and training family, model updating for structural digital twins. By learning equilibrium solves and encoding time integration, \modelname{} turns static knowledge into dynamic prediction without trajectory supervision.
\section{Methods}
\label{sec:method}
This section describes the formulation of \modelname{}. Additional theoretical and implementation details are provided in Supplementary Note~\ref{sec:supp-formulation}.

\subsection{Equilibrium formulation for implicit dynamics}
\label{subsec:equilibrium_formulation}

\modelname{} represents implicit dynamics as a rollout of effective equilibrium solves, enabling dynamic simulation through a solution map learned from static data (Fig.~\ref{fig:method}). We formulate this equilibrium representation from the spatially discretized equations of motion \cite{nonlinear-elasticity}:
\begin{equation}
    \mathbf{M}\ddot{\mathbf{u}}(t)
    + \mathbf{C}\dot{\mathbf{u}}(t)
    + \mathbf{r}_0\bigl(\mathbf{u}(t)\bigr)
    = \mathbf{f}_{\mathrm{ext}}(t),
    \label{eq:equation_of_motion_general}
\end{equation}
where $\mathbf{u}$, $\dot{\mathbf{u}}$ and $\ddot{\mathbf{u}}$ denote displacement, velocity, and acceleration, respectively. The matrices $\mathbf{M}$ and $\mathbf{C}$ are the mass and viscous-damping matrices, $\mathbf{r}_0(\mathbf{u})$ is the internal restoring-force vector, and $\mathbf{f}_{\mathrm{ext}}$ is the prescribed external force. The matrices $\mathbf{M}$ and $\mathbf{C}$ are constant, and the internal restoring force depends only on the current displacement. The equations are written over the free degrees of freedom after imposing the displacement boundary conditions. Linear elasticity is recovered when $\mathbf{r}_0(\mathbf{u})=\mathbf{K}_0\mathbf{u}$, whereas nonlinear elasticity is represented by a nonlinear dependence of
$\mathbf{r}_0$ on $\mathbf{u}$.

Given the state $\mathbf{s}^{n}=(\mathbf{u}^{n},\dot{\mathbf{u}}^{n},\ddot{\mathbf{u}}^{n})$, Newmark-$\beta$ integration \cite{newmark} expresses the next velocity and acceleration in terms of the unknown displacement $\mathbf{u}^{n+1}$ and the known state. For a time increment $\Delta t>0$ and parameters $(\beta,\gamma)$, with $\beta>0$, define
\begin{equation}
\begin{aligned}
    a_0 &= \frac{1}{\beta\Delta t^2},
    & a_1 &= \frac{\gamma}{\beta\Delta t},
    & a_2 &= \frac{1}{\beta\Delta t},
    \\
    a_3 &= \frac{1}{2\beta}-1,
    & a_4 &= \frac{\gamma}{\beta}-1,
    & a_5 &= \Delta t\left(\frac{\gamma}{2\beta}-1\right).
\end{aligned}
\label{eq:newmark_coefficients}
\end{equation}
Substituting the Newmark kinematic relations into Eq.~\eqref{eq:equation_of_motion_general} gives the effective equilibrium equation
\begin{equation}
    \mathbf{r}_0(\mathbf{u}^{n+1})
    + \mathbf{K}_{\mathrm{add}}^{\mathrm{dyn}}\mathbf{u}^{n+1}
    = \mathbf{F}_{\mathrm{eff}}^{n+1},
    \label{eq:effective_equilibrium}
\end{equation}
where the added-stiffness matrix and effective force are
\begin{equation}
\begin{aligned}
    \mathbf{K}_{\mathrm{add}}^{\mathrm{dyn}}
    &= a_0\mathbf{M}+a_1\mathbf{C},
    \\
    \mathbf{F}_{\mathrm{eff}}^{n+1}
    &= \mathbf{f}_{\mathrm{ext}}(t^{n+1})
    + \mathbf{M}\left(
        a_0\mathbf{u}^{n}
        + a_2\dot{\mathbf{u}}^{n}
        + a_3\ddot{\mathbf{u}}^{n}
      \right)
    + \mathbf{C}\left(
        a_1\mathbf{u}^{n}
        + a_4\dot{\mathbf{u}}^{n}
        + a_5\ddot{\mathbf{u}}^{n}
      \right).
\end{aligned}
\label{eq:newmark_assembly}
\end{equation}
Here, $t^{n+1}=t^{n}+\Delta t$, and we write $\mathbf{f}_{\mathrm{ext}}^{n+1}=\mathbf{f}_{\mathrm{ext}}(t^{n+1})$. Thus, the displacement at the next time step is determined by an equilibrium problem. The time increment determines the Newmark coefficients in both expressions, while the previous dynamic state enters through the effective force.

To connect this implicit step to static pretraining, consider equilibrium problems for the same system with added nodal springs:
\begin{equation}
    \mathbf{r}_0(\mathbf{u}^{\mathrm{sta}})
    + \mathbf{K}_{\mathrm{add}}^{\mathrm{sta}}\mathbf{u}^{\mathrm{sta}}
    = \mathbf{F}^{\mathrm{sta}},
    \qquad
    \mathbf{K}_{\mathrm{add}}^{\mathrm{sta}}
    = \mathbf{K}_{\mathrm{spring}}.
    \label{eq:static_equilibrium_general}
\end{equation}
Equations~\eqref{eq:effective_equilibrium} and~\eqref{eq:static_equilibrium_general} share the same equilibrium form when the internal-force law, boundary conditions, and added-stiffness representation are consistent. For lumped mass and mass-proportional damping, $\mathbf{C}=\alpha_M\mathbf{M}$, the dynamic added stiffness is diagonal and can be represented by nodal springs with coefficients $(a_0+a_1\alpha_M)m_i$, where $m_i$ is the lumped mass associated with degree of freedom $i$.

This formulation enables static-to-dynamic operator reuse in \modelname{}: a neural solver trained on static equilibrium data approximates the displacement solution at each implicit time step. Under the stated assumptions, the formulation accommodates general second-order elastodynamic systems through the restoring-force law $\mathbf{r}_0(\mathbf{u})$, while retaining the same effective-equilibrium form and Newmark kinematic relations. Further theoretical details are given in Supplementary Note~\ref{subsec:supp-physical-models}.

\subsection{\modelname{} architecture}
\label{subsec:architecture}

\modelname{} embeds a learned equilibrium solution map within a prescribed Newmark time-integration loop (Fig.~\ref{fig:method}). The architecture comprises effective-system construction, a neural equilibrium solve, and a kinematic state update. Only the equilibrium solution map is learned. The effective-system construction and state update follow the prescribed integration scheme.

Given the current rollout state $\mathbf{s}^{n}=(\mathbf{u}^{n},\dot{\mathbf{u}}^{n},\ddot{\mathbf{u}}^{n})$, initialized from the prescribed state $\mathbf{s}^{0}$, the next state is predicted as
\begin{equation}
    \mathbf{s}^{n+1}
    = \mathcal{U}_{\mathrm{NM}}\left(
        \mathbf{s}^{n},
        \mathcal{G}_{\theta}\left(
            \mathcal{F}_{\mathrm{NM}}\left(
                \mathbf{s}^{n},
                \mathbf{f}_{\mathrm{ext}}^{n+1}
            \right);\mathbf{X}
        \right)
    \right),
\label{eq:modelname_operator_composition}
\end{equation}
where $\mathcal{F}_{\mathrm{NM}}$ constructs the effective equilibrium inputs, $\mathcal{G}_{\theta}$ predicts displacement, and $\mathcal{U}_{\mathrm{NM}}$ updates the state while retaining the previous state as an explicit argument. The spatial coordinates are denoted by $\mathbf{X}$. Dependence on the prescribed $\mathbf{M}$, $\mathbf{C}$, $\Delta t$, $\beta$, and $\gamma$ is suppressed for readability.

\bmhead{Effective-system construction}
At each time step, $\mathcal{F}_{\mathrm{NM}}$ evaluates Eq.~\eqref{eq:newmark_assembly} using the current predicted state and the next external force. Its output is the pair $(\mathbf{k}_{\mathrm{add}}^{\mathrm{dyn}},\mathbf{F}_{\mathrm{eff}}^{n+1})$. For the lumped-mass and mass-proportional damping setting in Section~\ref{subsec:equilibrium_formulation}, the added-stiffness coefficients are computed directly as $k_{\mathrm{add},i}^{\mathrm{dyn}}=a_0m_i+a_1c_i$, with $c_i=\alpha_Mm_i$. Here, $m_i$ and $c_i$ denote the mass and damping coefficients associated with nodal degree of freedom $i$. This construction requires neither assembling an elastic stiffness matrix nor extracting its diagonal. The nodal coefficients represent the inertial and damping contributions to the effective equilibrium, while the baseline elastic response is represented by the learned solution map. The time increment enters both effective inputs through the Newmark coefficients, while the previous state enters through $\mathbf{F}_{\mathrm{eff}}^{n+1}$.

\bmhead{Neural equilibrium solve}
The neural solver maps these effective inputs to the next displacement field:
\begin{equation}
    \widehat{\mathbf{u}}^{n+1}
    = \mathcal{G}_{\theta}\left(
        \mathbf{k}_{\mathrm{add}}^{\mathrm{dyn}},
        \mathbf{F}_{\mathrm{eff}}^{n+1};\mathbf{X}
    \right).
    \label{eq:neural_solve}
\end{equation}
For each fixed structural system, the baseline restoring-force law and displacement constraints are represented through the static training data rather than supplied as a full stiffness matrix. The force and added-stiffness fields specify the equilibrium instance, while the coordinates provide spatial information.

\modelname{} can use any neural-operator backbone capable of representing the equilibrium solution map on the chosen spatial discretization. Here, we instantiate $\mathcal{G}_{\theta}$ using Transolver \cite{transolver} for irregular geometries and Fourier neural operator (FNO) \cite{fno} for regular nodal grids. Transolver receives fixed nodal coordinates alongside the force and added-stiffness fields, whereas FNO supplies grid coordinates internally as positional channels. In both cases, $\mathbf{X}$ denotes spatial information rather than trainable parameters. Embedding the learned equilibrium map within the prescribed time integration scheme enables zero-shot dynamic rollout from the neural operator representation.

\bmhead{Newmark state update}
Given the predicted displacement $\widehat{\mathbf{u}}^{n+1}$, $\mathcal{U}_{\mathrm{NM}}$ recovers the acceleration and velocity through the Newmark update relations:
\begin{equation}
\begin{aligned}
    \widehat{\ddot{\mathbf{u}}}^{n+1}
    &= a_0\left(\widehat{\mathbf{u}}^{n+1}-\widehat{\mathbf{u}}^{n}\right)
       -a_2\widehat{\dot{\mathbf{u}}}^{n}
       -a_3\widehat{\ddot{\mathbf{u}}}^{n},
    \\
    \widehat{\dot{\mathbf{u}}}^{n+1}
    &= \widehat{\dot{\mathbf{u}}}^{n}
       +\Delta t\left[
           (1-\gamma)\widehat{\ddot{\mathbf{u}}}^{n}
           +\gamma\widehat{\ddot{\mathbf{u}}}^{n+1}
       \right].
\end{aligned}
\label{eq:newmark_update}
\end{equation}
The updated state supplies the effective force at the next step. Iterating this produces a dynamic simulation. This composition establishes \modelname{} as a static-to-dynamic neural solver that advances the full dynamic state through a learned static equilibrium representation and prescribed time integration.

Each time step requires one neural equilibrium evaluation and effective-input and state updates prescribed by Newmark-$\beta$ integration, providing the rollout inference cost linear in the number of time steps for a fixed spatial discretization and network architecture. No stiffness-matrix factorization or Newton iteration is required during neural rollout. Rollout accuracy depends on the learned equilibrium approximation and the propagation of its errors through the time-integration scheme.

\subsection{Static pretraining and coverage design}
\label{subsec:pretrain}

Static pretraining learns the equilibrium solution map from static displacement labels. For each structural system, the training family retains the baseline restoring-force law, geometry, and displacement constraints while varying the applied static force. Added nodal stiffness is either fixed or varied according to the benchmark design. Each reference displacement satisfies Eq.~\eqref{eq:static_equilibrium_general}, giving the dataset
\begin{equation}
    \mathcal{D}_{\mathrm{sta}}
    = \left\{
        \left(
            \mathbf{k}_{\mathrm{add},j}^{\mathrm{sta}},
            \mathbf{F}_{j}^{\mathrm{sta}},
            \mathbf{u}_{j}^{\mathrm{sta}}
        \right)
    \right\}_{j=1}^{N_{\mathrm{sta}}},
    \label{eq:static_dataset}
\end{equation}
where $N_{\mathrm{sta}}$ is the number of static samples and $\mathbf{k}_{\mathrm{add},j}^{\mathrm{sta}}$ contains the nodal spring coefficients prescribed for sample $j$. The corresponding prediction is
\begin{equation}
    \widehat{\mathbf{u}}_{j}^{\mathrm{sta}}
    = \mathcal{G}_{\theta}\left(
        \mathbf{k}_{\mathrm{add},j}^{\mathrm{sta}},
        \mathbf{F}_{j}^{\mathrm{sta}};\mathbf{X}
    \right).
    \label{eq:static_prediction}
\end{equation}
The spatial coordinates $\mathbf{X}$ are shared across samples for the fixed system. This interface has the same form as the dynamic query in Eq.~\eqref{eq:neural_solve}, with static force and spring coefficients taking the roles of effective force and dynamic added stiffness. The equilibrium interface is not specific to FEM: other spatial discretizations can be used when their mass, damping, restoring-force, and boundary-condition representations are compatible with the static training family. The parameters $\theta$ are obtained by minimizing the mean relative $\ell_2$ displacement error over $\mathcal{D}_{\mathrm{sta}}$, $\mathcal{L}(\theta)=N_{\mathrm{sta}}^{-1}\sum_{j=1}^{N_{\mathrm{sta}}}\|\hat{\mathbf{u}}^{\mathrm{sta}}_j-\mathbf{u}^{\mathrm{sta}}_j\|_2/\|\mathbf{u}^{\mathrm{sta}}_j\|_2$, using the Adam optimizer.

\bmhead{Coverage of effective inputs}
The static training is designed to cover the force and added-stiffness inputs encountered during dynamic rollout. Force coverage concerns $\mathbf{F}_{\mathrm{eff}}^{n+1}$ in Eq.~\eqref{eq:newmark_assembly}, including the displacement, velocity, and acceleration contributions from the preceding state, rather than the external force alone. Sampling therefore targets the magnitudes and spatial variation of the effective-force fields.

For lumped mass and mass-proportional damping, the target dynamic added stiffness at degree of freedom $i$ is
\begin{equation}
    k_{\mathrm{add},i}^{\mathrm{dyn}}(\Delta t)
    = \left(
        \frac{1}{\beta\Delta t^2}
        + \frac{\gamma\alpha_M}{\beta\Delta t}
      \right)m_i.
    \label{eq:stiffness_coverage_target}
\end{equation}
For variable-stiffness training, samples target the joint variation of effective force and added stiffness, including their dependence on the time increment. The reported nonlinear benchmarks instead use a fixed added-stiffness field matched to the rollout setting; generalization across added stiffness is not assessed in these experiments. The shared equilibrium form establishes compatibility, but input coverage alone does not guarantee prediction accuracy or rollout stability. Within a validated range, the trained solver can support repeated forward evaluations and, potentially, inverse design or parameter identification, provided that the relevant parameter variations are represented in the training family.

\newpage
\section*{Extended Data}

\newcolumntype{Y}{>{\centering\arraybackslash}X}

\begin{table}[h!]
    \centering
    \scriptsize
    \caption{\textbf{1D dynamic benchmark.} Relative $\ell_2$ error (\%) of trajectory-trained baselines and static-trained \modelname{} on held-out static equilibrium samples (Sta) and in trajectory-level rollouts under IID, OOD(F), and OOD(H) conditions; lower is better. Sta is not applicable to the trajectory-trained baseline.}

    \label{tab:neural-1d}
    \setlength{\tabcolsep}{4pt}
    \renewcommand{\arraystretch}{1.15}

    \begin{tabularx}{\linewidth}{
        @{}
        >{\raggedright\arraybackslash}p{2.8cm}
        >{\raggedright\arraybackslash}p{3.2cm}
        >{\raggedleft\arraybackslash}p{0.9cm}
        *{4}{>{\centering\arraybackslash}X}
        @{}
    }
        \toprule
        & & & \multicolumn{4}{c}{Rel.\ $\ell_2$ [\%]\ $\downarrow$} \\
        \cmidrule(l){4-7}
        \textbf{Model} & \textbf{Evolution} & \textbf{Params}
        & Sta & IID & OOD(F) & OOD(H) \\
        \midrule

        \multicolumn{7}{@{}l}{%
          \textbf{Dynamic trajectory learning:}\ \
          $\mathbf{F}_{\mathrm{ext}}^{1:T} \rightarrow \mathbf{U}^{1:T}$} \\
        \midrule
        \addlinespace[3pt]

        FNO & learned $(x,t)$ map & 2.41M & -- & 3.57 & \underline{6.34} & 35.96$^{*}$ \\ \addlinespace[3pt]
        LSTM & learned recurrence & 3.88M & -- & 2.90 & 13.58 & 10.24 \\
        \addlinespace[3pt]
        Transolver & learned autoregression & 1.13M & -- & \underline{1.92} & 14.22 & \underline{9.74} \\
        \addlinespace[3pt]

        \midrule
        \addlinespace[3pt]

        \multicolumn{7}{@{}l}{\textbf{Static equilibrium learning and implicit time integrator:} $\mathbf{F}_{\mathrm{eff}}^{n+1} \rightarrow \mathbf{u}^{n+1}$} \\
        \midrule
        \addlinespace[3pt]

        \textbf{INOS} (Transolver) & encoded recurrence & 202k & 0.05 & 0.45 & 12.19 & 4.13 \\ \addlinespace[3pt]
        \textbf{INOS} (FNO) & encoded recurrence & 198k & \textbf{0.03} & \textbf{0.36} & \textbf{3.26} & \textbf{1.22} \\

        \bottomrule
    \end{tabularx}

    \vspace{2pt}
    \scriptsize
    $^{*}$Windowed inference is used for OOD(H).
\end{table}

\begin{table}[h!]
    \centering
    \scriptsize
    \caption{\textbf{2D dynamic benchmark.} Relative $\ell_2$ error (\%) of trajectory-trained baselines and static-trained \modelname{} on held-out static equilibrium samples (Sta) and in trajectory-level rollouts under IID, OOD(F), and OOD(H) conditions; lower is better. Sta is not applicable to the trajectory-trained baseline.}

    \label{tab:neural-2d}
    \setlength{\tabcolsep}{4pt}
    \renewcommand{\arraystretch}{1.15}

    \begin{tabularx}{\linewidth}{
        @{}
        >{\raggedright\arraybackslash}p{2.8cm}
        >{\raggedright\arraybackslash}p{3.2cm}
        >{\raggedleft\arraybackslash}p{0.9cm}
        *{4}{>{\centering\arraybackslash}X}
        @{}
    }
        \toprule
        & & & \multicolumn{4}{c}{Rel.\ $\ell_2$ [\%]\ $\downarrow$} \\
        \cmidrule(l){4-7}
        \textbf{Model} & \textbf{Evolution} & \textbf{Params}
        & Sta & IID & OOD(F) & OOD(H) \\
        \midrule

        \multicolumn{7}{@{}l}{%
          \textbf{Dynamic trajectory learning:}\ \
          $\mathbf{F}_{\mathrm{ext}}^{1:T} \rightarrow \mathbf{U}^{1:T}$} \\
        \midrule
        \addlinespace[3pt]

        FNO & learned $(x,y,t)$ map & 18.89M & -- & 7.96 & 10.34 & 9.42$^{*}$  \\
        \addlinespace[3pt]

        ConvLSTM & learned recurrence & 1.05M & -- & \underline{3.33} & \underline{4.87} & 8.34 \\
        \addlinespace[3pt]

        Transolver & learned autoregression & 1.42M & -- & 6.14 & 10.06 & \underline{3.06} \\
        \addlinespace[3pt]

        \midrule
        \addlinespace[3pt]

        \multicolumn{7}{@{}l}{%
          \textbf{Static equilibrium learning and implicit time integrator:} $\mathbf{F}_{\mathrm{eff}}^{n+1} \rightarrow \mathbf{u}^{n+1}$} \\
        \midrule
        \addlinespace[3pt]

        \textbf{INOS} (Transolver) & encoded recurrence & 728k & 0.60 & 1.02 & \textbf{0.83} & 1.13 \\
        \addlinespace[3pt]

        \textbf{INOS} (FNO) & encoded recurrence & 603k & \textbf{0.53} & \textbf{0.69} & 0.91 & \textbf{0.74} \\

        \bottomrule
    \end{tabularx}

    \vspace{2pt}
    \scriptsize
    $^{*}$Windowed inference is used for OOD(H).
\end{table}

\newpage
\backmatter

\bmhead{Supplementary information}

Supplementary Information is available for this manuscript.

\bmhead{Acknowledgements}

C.K. acknowledges support from the Hong Kong PhD Fellowship Scheme of the Research Grants Council of Hong Kong.

\section*{Data availability}

The datasets generated and analyzed in this study will be made available upon publication. Data required to reproduce the results are available to editors and reviewers upon request during peer review.

\section*{Code availability}

The source code will be made available upon publication. 

\section*{Declarations}

\bmhead{Funding}
This work was supported by the Research Grants Council of the Hong Kong Special Administrative Region, China, through the Early Career Scheme (Grant No. 27201226) and the Innovation and Technology Fund of Hong Kong through the Innovation and Technology Support Programme (Grant No. ITS/041/23MX). Support was also provided by the National Natural Science Foundation of China through the Young Scientists Fund (Grant No. 52408221) and the Environment and Conservation Fund of Hong Kong (Grant No. ECF249/2024).

\bmhead{Competing interests}

The authors declare no competing interests.

\bmhead{Ethics approval and consent to participate}

Not applicable.

\bmhead{Consent for publication}

Not applicable.

\bmhead{Materials availability}

Not applicable.

\bmhead{Author contributions}

C.K.  and J.W. conceived and designed the study. X.N. contributed to the conceptualization of the study. C.K. developed the methodology, conducted the numerical experiments, analyzed the results, and prepared the figures. J.W. supervised the research and acquired funding. C.K. drafted the manuscript, and J.W. revised it. All authors reviewed and approved the final manuscript.

\subsection*{Use of generative AI}

The authors used ChatGPT (OpenAI) and Claude (Anthropic) to assist with revising and restructuring the manuscript text, checking the consistency of notation, terminology, and equations, and formatting the plotting scripts used for figures. All AI-assisted output was reviewed, edited, and verified by the authors, who take full responsibility for the content of this manuscript.

\backmatter

\bibliography{sn-bibliography}

\newpage
\begin{appendices}

\section*{Supplementary Note}

\begin{table}[h!]
\centering
\caption{\textbf{Overview of the Supplementary Information.}}
\label{tab:supplementary-overview}
\small
\setlength{\tabcolsep}{5pt}
\renewcommand{\arraystretch}{1.25}

\begin{tabularx}{\linewidth}{
    >{\raggedright\arraybackslash}p{0.05\linewidth}
    >{\raggedright\arraybackslash}p{0.25\linewidth}
    >{\raggedright\arraybackslash}X
}
\toprule
& \textbf{Title} & \textbf{Key content and purpose} \\
\midrule

\ref{sec:supp-formulation} &
\nameref{sec:supp-formulation}  &
\ref{subsec:supp-physical-models}: \nameref{subsec:supp-physical-models} \newline
\ref{subsec:supp-rollout}: \nameref{subsec:supp-rollout} \newline
\ref{subsec:supp-error-decomp}: \nameref{subsec:supp-error-decomp} \newline
\ref{subsec:supp-complexity}: \nameref{subsec:supp-complexity} \newline
\ref{subsec:supp-differentiability}: \nameref{subsec:supp-differentiability}
\\
\midrule

\ref{sec:supp-data-protocols} &
\nameref{sec:supp-data-protocols} &
\ref{subsec:supp-problem-configuration}: \nameref{subsec:supp-problem-configuration}\newline
\ref{subsec:supp-mesh-convergence}: \nameref{subsec:supp-mesh-convergence}\newline
\ref{subsec:supp-static-data}: \nameref{subsec:supp-static-data}\newline
\ref{subsec:supp-dynamic-regimes}: \nameref{subsec:supp-dynamic-regimes}\newline
\ref{subsec:supp-metrics}: \nameref{subsec:supp-metrics}\newline
\ref{subsec:supp-computational-environment}: \nameref{subsec:supp-computational-environment}
\\
\midrule

\ref{sec:supp-neural-methods} & \nameref{sec:supp-neural-methods} & \ref{subsec:supp-dynamic-baselines}:\nameref{subsec:supp-dynamic-baselines}\newline
\ref{subsec:supp-neural-configurations}: \nameref{subsec:supp-neural-configurations}\newline
\ref{subsec:supp-data-coverage}:\nameref{subsec:supp-data-coverage}
\\
\midrule

\ref{sec:supp-coverage} & \nameref{sec:supp-coverage} &
\ref{subsec:supp-coverage-condition}: \nameref{subsec:supp-coverage-condition}\newline
\ref{subsec:supp-coverage-experiment-design}: \nameref{subsec:supp-coverage-experiment-design} \newline
\ref{subsec:supp-coverage-result}: \nameref{subsec:supp-coverage-result}\newline
\ref{subsec:supp-million-step}: \nameref{subsec:supp-million-step}
\\
\midrule
\ref{sec:supp-resolution-transfer} & \nameref{sec:supp-resolution-transfer} & 
\\
\bottomrule
\end{tabularx}
\end{table}

\newpage
\section{Theoretical formulation}
\label{sec:supp-formulation}

\subsection{Elastodynamic theory and \modelname{} formulation}
\label{subsec:supp-physical-models}

We formulate linear elasticity and hyperelasticity through the restoring-force operator $\mathbf{r}_0(\mathbf{u})$ in Methods (Section~\ref{sec:method}), independently of a particular spatial discretization.

\subsubsection{Linear elasticity}
\label{subsubsec:supp-linear-elastic}
For small displacement gradients on a reference domain $\Omega$, the displacement field $\mathbf{u}(\mathbf{x})$ gives the infinitesimal strain and linear elastic Cauchy stress \cite{nonlinear-solid-mech},
\begin{equation}
\begin{aligned}
    \boldsymbol{\varepsilon}(\mathbf{u})
    &= \frac{1}{2}\left[
        \nabla\mathbf{u}+(\nabla\mathbf{u})^{\mathsf{T}}
       \right],
    \\
    \boldsymbol{\sigma}(\mathbf{u})
    &= \mathbb{C}:\boldsymbol{\varepsilon}(\mathbf{u}),
\end{aligned}
\label{eq:supp-linear-constitutive}
\end{equation}
where $\mathbb{C}$ is the elasticity tensor and the colon denotes tensor contraction. The strain energy is
\begin{equation}
    \mathcal{E}_0[\mathbf{u}]
    = \frac{1}{2}\int_{\Omega}
        \boldsymbol{\varepsilon}(\mathbf{u})
        :\mathbb{C}:\boldsymbol{\varepsilon}(\mathbf{u})
        \,\mathrm{d}\Omega.
\label{eq:supp-linear-energy}
\end{equation}
After energy-consistent spatial discretization with homogeneous displacement constraints, $\mathbf{u}$ denotes the free displacement vector. The discrete energy $\mathcal{E}_0(\mathbf{u})=\tfrac{1}{2}\mathbf{u}^{\mathsf{T}}\mathbf{K}_0\mathbf{u}$ gives the restoring force and constant tangent stiffness,
\begin{equation}
    \mathbf{r}_0(\mathbf{u})
    = \nabla_{\mathbf{u}}\mathcal{E}_0(\mathbf{u})
    = \mathbf{K}_0\mathbf{u},
    \qquad
    \frac{\partial\mathbf{r}_0}{\partial\mathbf{u}}
    = \mathbf{K}_0.
\label{eq:supp-linear-restoring-force}
\end{equation}
Equation~\eqref{eq:effective_equilibrium} therefore reduces to $(\mathbf{K}_0+\mathbf{K}_{\mathrm{add}}^{\mathrm{dyn}})\mathbf{u}^{n+1}=\mathbf{F}_{\mathrm{eff}}^{n+1}$.

\subsubsection{Nonlinear elasticity}
\label{subsubsec:supp-nonlinear-elastic}
For finite deformations, we use a hyperelastic energy density $W(\mathbf{F})$ per unit reference volume, twice differentiable over the admissible deformation range \cite{nonlinear-solid-mech}. With $\mathbf{X}$ denoting material coordinates, the deformation gradient and volume ratio are
\begin{equation}
    \mathbf{F}=\mathbf{I}+\nabla_{\mathbf{X}}\mathbf{u},
    \qquad
    J=\det\mathbf{F}>0,
\label{eq:supp-hyperelastic-kinematics}
\end{equation}
where $\mathbf{I}$ is the identity tensor. The elastic energy and first Piola-Kirchhoff stress are
\begin{equation}
    \mathcal{E}_0[\mathbf{u}]
    = \int_{\Omega}W(\mathbf{F})\,\mathrm{d}\Omega,
    \qquad
    \mathbf{P}=\frac{\partial W}{\partial\mathbf{F}}.
\label{eq:supp-hyperelastic-energy}
\end{equation}
In the same discrete setting, the energy $\mathcal{E}_0(\mathbf{u})$ gives
\begin{equation}
    \mathbf{r}_0(\mathbf{u})
    = \nabla_{\mathbf{u}}\mathcal{E}_0(\mathbf{u}),
    \qquad
    \mathbf{K}_{\mathrm{tan}}(\mathbf{u})
    = \frac{\partial\mathbf{r}_0}{\partial\mathbf{u}}
    = \nabla_{\mathbf{u}}^2\mathcal{E}_0(\mathbf{u}).
\label{eq:supp-hyperelastic-restoring-force}
\end{equation}
The tangent $\mathbf{K}_{\mathrm{tan}}$ generally varies with displacement, and the restoring force cannot be replaced by $\mathbf{K}_{\mathrm{tan}}(\mathbf{u})\mathbf{u}$. For constant mass and damping matrices, the Newmark residual and Jacobian are
\begin{equation}
\begin{aligned}
    \mathbf{R}^{n+1}(\mathbf{u})
    &= \mathbf{r}_0(\mathbf{u})
       +\mathbf{K}_{\mathrm{add}}^{\mathrm{dyn}}\mathbf{u}
       -\mathbf{F}_{\mathrm{eff}}^{n+1},
    \\
    \frac{\partial\mathbf{R}^{n+1}}{\partial\mathbf{u}}
    &= \mathbf{K}_{\mathrm{tan}}(\mathbf{u})
       +\mathbf{K}_{\mathrm{add}}^{\mathrm{dyn}}.
\end{aligned}
\label{eq:supp-hyperelastic-effective-residual}
\end{equation}
A fixed time step makes the added stiffness constant, but not generally the total tangent.

For both models, static pretraining and dynamic rollout use the same restoring-force law, displacement constraints, and compatible discretizations. In \modelname{}, Eq.~\eqref{eq:neural_solve} maps force and added-stiffness fields to displacement fields.

\subsection{Implementation details of \modelname{} rollout}
\label{subsec:supp-rollout}

Algorithm~\ref{alg:inos_rollout} implements the effective-equilibrium formulation in Eq.~\eqref{eq:effective_equilibrium} for both linear and nonlinear elasticity. We use constant lumped mass coefficients $\mathbf{m}$ over the free degrees of freedom and mass-proportional damping $\mathbf{C}=\alpha_M\mathbf{M}$. Matrix-vector products in the Newmark assembly are evaluated at each node as $\mathbf{M}\mathbf{v}=\mathbf{m}\odot\mathbf{v}$ and $\mathbf{C}\mathbf{v}=\alpha_M\mathbf{m}\odot\mathbf{v}$, where $\odot$ denotes component-wise multiplication. The baseline restoring-force law and displacement constraints are represented by the corresponding statically pretrained solver $\mathcal{G}_{\theta}$, whose inputs are the added-stiffness and effective-force fields, together with spatial coordinates $\mathbf{X}$, as in Eq.~\eqref{eq:neural_solve}.

The rollout starts from a consistent initial state $\mathbf{s}^{0}=(\mathbf{u}^{0},\dot{\mathbf{u}}^{0},\ddot{\mathbf{u}}^{0})$. Given initial displacement and velocity, the initial acceleration is obtained before rollout from
\begin{equation}
    \mathbf{m}\odot\ddot{\mathbf{u}}^{0}
    = \mathbf{f}_{\mathrm{ext}}(t^{0})
      -\alpha_M\mathbf{m}\odot\dot{\mathbf{u}}^{0}
      -\mathbf{r}_0(\mathbf{u}^{0}),
\label{eq:supp-rollout-initial-acceleration}
\end{equation}
with positive mass coefficients on the free degrees of freedom. A supplied initial acceleration must satisfy the same balance.

We write the algorithm for prescribed positive increments $\{\Delta t_n\}_{n=0}^{T-1}$, with $t^{n+1}=t^n+\Delta t_n$ and $T$ denoting the number of steps. Fixed-step rollout is the special case $\Delta t_n=\Delta t$. In that case, the Newmark coefficients and added-stiffness field can be computed once before the loop and reused. However, this does not imply a constant nonlinear tangent stiffness. The nonlinear tangent stiffness remains displacement-dependent and varies from one time step to the next.

\begin{algorithm}[h!]
\caption{\modelname{} rollout with fixed or prescribed variable time steps}
\label{alg:inos_rollout}
\begin{algorithmic}[1]
\Require Initial state $\mathbf{s}^{0}$ at $t^0$; external-load history $\mathbf{f}_{\mathrm{ext}}(t)$; lumped mass $\mathbf{m}$; damping coefficient $\alpha_M$; coordinates $\mathbf{X}$; prescribed increments $\{\Delta t_n\}_{n=0}^{T-1}$; Newmark parameters $(\beta,\gamma)$ with $\beta>0$; pretrained solver $\mathcal{G}_{\theta}$

\State Initialize $\widehat{\mathbf{s}}^{0}\leftarrow\mathbf{s}^{0}$
\For{$n=0,\ldots,T-1$}
    \State Set $t^{n+1}\leftarrow t^n+\Delta t_n$
    \State Compute $a_0^n,\ldots,a_5^n$ from Eq.~\eqref{eq:newmark_coefficients} using $\Delta t_n$
    \State Compute $\mathbf{k}_{\mathrm{add}}^{\mathrm{dyn},n}\leftarrow(a_0^n+a_1^n\alpha_M)\mathbf{m}$
    \State Assemble $\mathbf{F}_{\mathrm{eff}}^{n+1}$ using Eq.~\eqref{eq:newmark_assembly}, the predicted state $\widehat{\mathbf{s}}^n$, coefficients $a_0^n,\ldots,a_5^n$, and nodal mass and damping products
    \State Predict $\widehat{\mathbf{u}}^{n+1}\leftarrow\mathcal{G}_{\theta}\bigl(\mathbf{k}_{\mathrm{add}}^{\mathrm{dyn},n},\mathbf{F}_{\mathrm{eff}}^{n+1};\mathbf{X}\bigr)$
    \State Recover $\widehat{\ddot{\mathbf{u}}}^{n+1}\leftarrow a_0^n\bigl(\widehat{\mathbf{u}}^{n+1}-\widehat{\mathbf{u}}^n\bigr)-a_2^n\widehat{\dot{\mathbf{u}}}^{n}-a_3^n\widehat{\ddot{\mathbf{u}}}^{n}$
    \State Recover $\widehat{\dot{\mathbf{u}}}^{n+1}\leftarrow\widehat{\dot{\mathbf{u}}}^{n}+\Delta t_n\bigl[(1-\gamma)\widehat{\ddot{\mathbf{u}}}^{n}+\gamma\widehat{\ddot{\mathbf{u}}}^{n+1}\bigr]$
    \State Set $\widehat{\mathbf{s}}^{n+1}\leftarrow\bigl(\widehat{\mathbf{u}}^{n+1},\widehat{\dot{\mathbf{u}}}^{n+1},\widehat{\ddot{\mathbf{u}}}^{n+1}\bigr)$
\EndFor
\end{algorithmic}
\end{algorithm}

\bmhead{Variable time stepping}
Changing $\Delta t_n$ updates both $\mathbf{k}_{\mathrm{add}}^{\mathrm{dyn},n}$ and $\mathbf{F}_{\mathrm{eff}}^{n+1}$. \modelname{} supports prescribed variable stepping and could further support adaptive stepping. Linear-elastic training covers multiple increments, whereas nonlinear-elastic training uses a single parameter $\kappa=a_0+a_1\alpha_M$ matched to the fixed rollout increment. Variable-step nonlinear rollout is therefore a direct extension, provided that training covers the corresponding added-stiffness and effective-force ranges.

\subsection{Error decomposition in \modelname{}}
\label{subsec:supp-error-decomp}

We compare the learned rollout with the Newmark rollout using exact equilibrium solves, under the same spatial discretization, loading, initial state, and prescribed time increments. This isolates the surrogate error and excludes spatial and temporal discretization errors. All quantities are restricted to the free degrees of freedom.

\bmhead{Local solve error}

For an input $\mathbf{z}=(\mathbf{k},\mathbf{F})$, let $\mathcal{S}(\mathbf{z})$ denote the unique admissible equilibrium satisfying
\begin{equation}
    \mathbf{r}_0(\mathbf{u})
    +\operatorname{Diag}(\mathbf{k})\mathbf{u}
    =\mathbf{F}.
\label{eq:supp-equilibrium-solution-operator}
\end{equation}
Existence and uniqueness are assumed over the input regime considered. At step $n$, write
\begin{equation}
    \mathbf{K}_{\mathrm{add},n}
    =\operatorname{Diag}
      \bigl(\mathbf{k}_{\mathrm{add}}^{\mathrm{dyn},n}\bigr)
    =\kappa_n\mathbf{M},
    \qquad
    \kappa_n=a_0^n+a_1^n\alpha_M>0.
\label{eq:supp-added-stiffness-matrix}
\end{equation}
For the input $\widehat{\mathbf{z}}^{n+1}$ assembled from
$\widehat{\mathbf{s}}^n$, define
\begin{equation}
\begin{aligned}
    \widetilde{\mathbf{u}}^{n+1}
    &=\mathcal{S}(\widehat{\mathbf{z}}^{n+1}),\\
    \boldsymbol{\delta}^{n+1}
    &=\widehat{\mathbf{u}}^{n+1}
      -\widetilde{\mathbf{u}}^{n+1},\\
    \boldsymbol{\rho}^{n+1}
    &=\mathbf{r}_0(\widehat{\mathbf{u}}^{n+1})
      +\mathbf{K}_{\mathrm{add},n}\widehat{\mathbf{u}}^{n+1}
      -\widehat{\mathbf{F}}_{\mathrm{eff}}^{n+1}.
\end{aligned}
\label{eq:supp-local-error-and-residual}
\end{equation}
Here, $\widehat{\mathbf{u}}^{n+1}
=\mathcal{G}_{\theta}(\widehat{\mathbf{z}}^{n+1};\mathbf{X})$.
The local error $\boldsymbol{\delta}^{n+1}$ compares learned and exact solves from the same predicted previous state. By the Newmark recovery relations, the residual also equals the dynamic equilibrium defect:
\begin{equation}
\begin{aligned}
    \boldsymbol{\rho}^{n+1}
    ={}&
    \mathbf{M}\widehat{\ddot{\mathbf{u}}}^{n+1}
    +\mathbf{C}\widehat{\dot{\mathbf{u}}}^{n+1}
    +\mathbf{r}_0(\widehat{\mathbf{u}}^{n+1})
    -\mathbf{f}_{\mathrm{ext}}(t^{n+1}).
\end{aligned}
\label{eq:supp-dynamic-equilibrium-residual}
\end{equation}
Assume that $\mathbf{r}_0$ is continuously differentiable on a convex displacement region containing the compared solutions and predictions, and that
\begin{equation}
    \mathbf{J}_n(\mathbf{u})
    =D\mathbf{r}_0(\mathbf{u})+\mathbf{K}_{\mathrm{add},n},
    \qquad
    \operatorname{sym}\mathbf{J}_n(\mathbf{u})
    \succeq\mu_n\mathbf{I},
    \qquad \mu_n>0,
\label{eq:supp-effective-tangent-bound}
\end{equation}
where $\operatorname{sym}\mathbf{J}=(\mathbf{J}+\mathbf{J}^{\mathsf{T}})/2$. Define the effective restoring-force map and the displacement path
\begin{equation}
\begin{aligned}
    \mathbf{H}_n(\mathbf{u})
    &=
    \mathbf{r}_0(\mathbf{u})
    +\mathbf{K}_{\mathrm{add},n}\mathbf{u},\\
    \mathbf{u}(\xi)
    &=
    \widetilde{\mathbf{u}}^{n+1}
    +\xi\boldsymbol{\delta}^{n+1},
    \qquad \xi\in[0,1].
\end{aligned}
\label{eq:supp-effective-map-and-path}
\end{equation}
Along this path, the chain rule gives
\begin{equation}
    \frac{\mathrm{d}}{\mathrm{d}\xi}
    \mathbf{H}_n(\mathbf{u}(\xi))
    =
    \mathbf{J}_n(\mathbf{u}(\xi))
    \boldsymbol{\delta}^{n+1}.
\label{eq:supp-path-chain-rule}
\end{equation}
Integrating Eq.~\eqref{eq:supp-path-chain-rule} from $\xi=0$ to $\xi=1$ and using the equilibrium equation for $\widetilde{\mathbf{u}}^{n+1}$ yields
\begin{equation}
\begin{aligned}
    \boldsymbol{\rho}^{n+1} &= \mathbf{H}_n(\mathbf{u}(1)) - \mathbf{H}_n(\mathbf{u}(0))\\
    &= \int_0^1 \frac{\mathrm{d}}{\mathrm{d}\xi} \mathbf{H}_n(\mathbf{u}(\xi)) \,\mathrm{d}\xi\\
    &=\left[ \int_0^1 \mathbf{J}_n \bigl( \widetilde{\mathbf{u}}^{n+1} +\xi\boldsymbol{\delta}^{n+1} \bigr) \,\mathrm{d}\xi \right]\boldsymbol{\delta}^{n+1}.
\end{aligned}
\label{eq:supp-residual-mean-value}
\end{equation}
Taking the inner product of Eq.~\eqref{eq:supp-residual-mean-value} with $\boldsymbol{\delta}^{n+1}$ and applying Eq.~\eqref{eq:supp-effective-tangent-bound} and the Cauchy-Schwarz inequality gives
\begin{equation}
    \mu_n\|\boldsymbol{\delta}^{n+1}\|_2^2 \leq (\boldsymbol{\delta}^{n+1})^{\mathsf{T}} \boldsymbol{\rho}^{n+1} \leq \|\boldsymbol{\delta}^{n+1}\|_2 \|\boldsymbol{\rho}^{n+1}\|_2.
\label{eq:supp-residual-inner-product-bound}
\end{equation}
Dividing by $\|\boldsymbol{\delta}^{n+1}\|_2$ when the local error is nonzero yields
\begin{equation}
    \|\boldsymbol{\delta}^{n+1}\|_2
    \leq
    \frac{\|\boldsymbol{\rho}^{n+1}\|_2}{\mu_n}.
\label{eq:supp-residual-error-bound}
\end{equation}
The bound also holds when the local error is zero. Thus, the same coercivity condition ensures regional uniqueness and controls the local solve error.

For linear elasticity with $\mathbf{K}_0\succeq0$, a directly computable choice is $\mu_n=\kappa_n m_{\min}$, where $m_{\min}=\min_i m_i>0$. For nonlinear elasticity, if $\operatorname{sym}D\mathbf{r}_0(\mathbf{u})\succeq-c_n\mathbf{I}$ throughout the comparison region, one may take 
\begin{equation}
    \mu_n=\kappa_n m_{\min}-c_n>0.
\label{eq:supp-nonlinear-coercivity-bound}
\end{equation}
On a fixed region with bounded negative curvature, sufficiently small time increments make this bound positive. Evaluating the residual requires one restoring-force evaluation and no reference trajectory; using it as an error certificate additionally requires a valid positive lower bound $\mu_n$.

To relate the local error to static pretraining, let
$\mathcal{D}_{\mathrm{sta}}$ denote the joint support of the added-stiffness and force inputs, and assume
\begin{equation}
    \sup_{\mathbf{z}\in\mathcal{D}_{\mathrm{sta}}}
    \left\|
        \mathcal{G}_{\theta}(\mathbf{z};\mathbf{X})
        -\mathcal{S}(\mathbf{z})
    \right\|_2
    \leq\varepsilon_{\theta}.
\label{eq:supp-static-support-error}
\end{equation}
Suppose that $\mathcal{G}_{\theta}$ and $\mathcal{S}$ have uniform Lipschitz constants $L_G$ and $L_S$ on a common input domain containing the static support and the rollout inputs. Let $\|\cdot\|_{\mathcal{Z}}$ denote the fixed scaled product norm used for input distances. For any $\mathbf{z}_s\in\mathcal{D}_{\mathrm{sta}}$, adding and subtracting the learned and exact solutions at $\mathbf{z}_s$ and applying the triangle inequality gives
\begin{equation}
\begin{aligned}
    \|\boldsymbol{\delta}^{n+1}\|_2
    &\leq
    \left\|
        \mathcal{G}_{\theta}
        (\widehat{\mathbf{z}}^{n+1};\mathbf{X})
        -\mathcal{G}_{\theta}(\mathbf{z}_s;\mathbf{X})
    \right\|_2+
    \left\|
        \mathcal{G}_{\theta}(\mathbf{z}_s;\mathbf{X})
        -\mathcal{S}(\mathbf{z}_s)
    \right\|_2+
    \left\|
        \mathcal{S}(\mathbf{z}_s)
        -\mathcal{S}(\widehat{\mathbf{z}}^{n+1})
    \right\|_2\\
    &\leq
    \varepsilon_{\theta}
    +(L_G+L_S)
    \|\widehat{\mathbf{z}}^{n+1}
      -\mathbf{z}_s\|_{\mathcal{Z}}.
\end{aligned}
\label{eq:supp-coverage-triangle-bound}
\end{equation}
Taking the infimum over $\mathbf{z}_s\in\mathcal{D}_{\mathrm{sta}}$ yields
\begin{equation}
\begin{aligned}
    \|\boldsymbol{\delta}^{n+1}\|_2
    \leq{}&
    \varepsilon_{\theta}
    +(L_G+L_S)
    \operatorname{dist}
    \bigl(
        \widehat{\mathbf{z}}^{n+1},
        \mathcal{D}_{\mathrm{sta}}
    \bigr)
    =:d_n.
\end{aligned}
\label{eq:supp-input-coverage-bound}
\end{equation}
The two terms describe approximation error on the static support and joint input-coverage mismatch. Both contribute to the same local defect. The residual bound in Eq.~\eqref{eq:supp-residual-error-bound} may replace $d_n$ whenever it is smaller. No separate fixed-stiffness operator mismatch arises because the learned and exact solves use the same added-stiffness input.

The per-step error is therefore bounded in terms of static approximation and joint input coverage, and the computable residual provides a local error bound whenever a valid $\mu_n>0$ is available.

\bmhead{Rollout error} 
Let $\mathbf{s}_{\star}^n$ denote the reference state, $\mathbf{e}^n=\widehat{\mathbf{s}}^n-\mathbf{s}_{\star}^n$, and $\Phi_n$ the exact Newmark state map. The recovery relations give the predicted-state identity, and subtracting the reference update gives the state-error recurrence
\begin{equation}
\begin{aligned}
    \widehat{\mathbf{s}}^{n+1}
    &=
    \Phi_n(\widehat{\mathbf{s}}^n)
    +\mathbf{B}_n\boldsymbol{\delta}^{n+1},
    \qquad \text{with}\quad
    \mathbf{B}_n=
    \begin{bmatrix}
        \mathbf{I}\\
        a_1^n\mathbf{I}\\
        a_0^n\mathbf{I}
    \end{bmatrix},\\
    \mathbf{e}^{n+1}
    &=
    \widehat{\mathbf{s}}^{n+1}
    -\mathbf{s}_{\star}^{n+1}\\
    &=
    \Phi_n(\widehat{\mathbf{s}}^n)
    -\Phi_n(\mathbf{s}_{\star}^n)
    +\mathbf{B}_n\boldsymbol{\delta}^{n+1}.
\end{aligned}
\label{eq:supp-state-error-injection}
\end{equation}
The error also admits an exact interpretation as the response
to the residual load. Let $\mathcal{A}_n$ denote the linear
part of the effective-force assembly, and define the averaged
tangent
\begin{equation}
    \overline{\mathbf{K}}_T^{\,n+1}
    =
    \int_0^1
    D\mathbf{r}_0
    \bigl(
        \mathbf{u}_{\star}^{n+1}
        +\xi\mathbf{e}_u^{n+1}
    \bigr)
    \,\mathrm{d}\xi.
\label{eq:supp-averaged-tangent}
\end{equation}
Subtracting the reference effective-equilibrium equation from the learned equation and applying the integral mean-value identity gives
\begin{equation}
\begin{aligned}
    \mathcal{A}_n\mathbf{e}^n+\boldsymbol{\rho}^{n+1}
    &=
    \widehat{\mathbf{F}}_{\mathrm{eff}}^{n+1}
    -\mathbf{F}_{\mathrm{eff},\star}^{n+1}
    +\boldsymbol{\rho}^{n+1}\\
    &=
    \mathbf{r}_0(\widehat{\mathbf{u}}^{n+1})
    -\mathbf{r}_0(\mathbf{u}_{\star}^{n+1})
    + \mathbf{K}_{\mathrm{add},n}
    \bigl(
        \widehat{\mathbf{u}}^{n+1}
        -\mathbf{u}_{\star}^{n+1}
    \bigr)\\
    &=
    \bigl(
        \overline{\mathbf{K}}_T^{\,n+1}
        +\mathbf{K}_{\mathrm{add},n}
    \bigr)\mathbf{e}_u^{n+1}.
\end{aligned}
\label{eq:supp-error-effective-equilibrium}
\end{equation}
The state differences satisfy the Newmark recovery relations. Subtracting the reference dynamic-equilibrium equation from Eq.~\eqref{eq:supp-dynamic-equilibrium-residual} and using Eq.~\eqref{eq:supp-averaged-tangent} gives
\begin{equation}
\begin{aligned}
    \boldsymbol{\rho}^{n+1}
    &=
    \mathbf{M}
    \bigl(
        \widehat{\ddot{\mathbf{u}}}^{n+1}
        -\ddot{\mathbf{u}}_{\star}^{n+1}
    \bigr)
    +\mathbf{C}
    \bigl(
        \widehat{\dot{\mathbf{u}}}^{n+1}
        -\dot{\mathbf{u}}_{\star}^{n+1}
    \bigr) 
    + \mathbf{r}_0(\widehat{\mathbf{u}}^{n+1})
    -\mathbf{r}_0(\mathbf{u}_{\star}^{n+1})\\
    &=
    \mathbf{M}\mathbf{e}_a^{n+1}
    +\mathbf{C}\mathbf{e}_v^{n+1}
    +\overline{\mathbf{K}}_T^{\,n+1}\mathbf{e}_u^{n+1}.
\end{aligned}
\label{eq:supp-residual-load-error-dynamics}
\end{equation}
Thus, the error is a Newmark response to the residual load with zero initial error. This identity requires no linearization. For linear elasticity, $\overline{\mathbf{K}}_T^{\,n+1}=\mathbf{K}_0$. For nonlinear elasticity, the averaged tangent depends on both trajectories and generally varies with time; the identity does not imply unconditional stability.

For a full-state bound, introduce a fixed reference time $\tau>0$ and the dimensionally consistent norm
\begin{equation}
    \|\mathbf{e}\|_{\tau}^2
    =
    \|\mathbf{e}_u\|_2^2
    +\tau^2\|\mathbf{e}_v\|_2^2
    +\tau^4\|\mathbf{e}_a\|_2^2.
\label{eq:supp-scaled-state-norm}
\end{equation}
Suppose that $\Phi_n$ has Lipschitz constant $\ell_n$ on the compared states in the norm $\|\cdot\|_{\tau}$. Using Eq.~\eqref{eq:supp-state-error-injection} and the local-error bound in Eq.~\eqref{eq:supp-input-coverage-bound} gives
\begin{equation}
\begin{aligned}
    \|\mathbf{e}^{n+1}\|_{\tau}
    &\leq
    \left\|
        \Phi_n(\widehat{\mathbf{s}}^n)
        -\Phi_n(\mathbf{s}_{\star}^n)
    \right\|_{\tau}
    +\|\mathbf{B}_n\boldsymbol{\delta}^{n+1}\|_{\tau}\\
    &\leq
    \ell_n\|\mathbf{e}^n\|_{\tau}
    +b_n\|\boldsymbol{\delta}^{n+1}\|_2\\
    &\leq
    \ell_n\|\mathbf{e}^n\|_{\tau}+b_nd_n,
\end{aligned}
\label{eq:supp-one-step-error-decomposition}
\end{equation}
where
\begin{equation}
    b_n=\|\mathbf{B}_n\|_{2\to\tau} =\sqrt{1+\tau^2(a_1^n)^2+\tau^4(a_0^n)^2}.
\end{equation}
Define the scalar state error $E_n:=\|\mathbf{e}^n\|_{\tau}$, with $E_0=0$. Repeated substitution of Eq.~\eqref{eq:supp-one-step-error-decomposition} yields
\begin{equation}
\begin{aligned}
    E_n
    &\leq
    \ell_{n-1}E_{n-1}+b_{n-1}d_{n-1}\\
    &\leq
    E_0\prod_{q=0}^{n-1}\ell_q
    +
    \sum_{j=0}^{n-1}
    b_jd_j
    \prod_{q=j+1}^{n-1}\ell_q\\
    &=
    \sum_{j=0}^{n-1}
    b_jd_j
    \prod_{q=j+1}^{n-1}\ell_q,
\end{aligned}
\label{eq:supp-finite-horizon-error-bound}
\end{equation}
where an empty product equals one. The step map need not be a contraction in this norm, so the product can grow exponentially with the horizon. This is a finite-horizon Gronwall-type estimate, conditional on the trajectories remaining within the stated comparison regions.

The learned solver therefore enters the rollout error only through the residual loads it introduces, which propagate through the Newmark error dynamics of the structure rather than a learned temporal map.

\bmhead{Bounded accumulation}

For linear elasticity with a fixed increment, let $\mathbf{P}$ denote the exact Newmark amplification matrix and $\mathbf{K}_{\mathrm{eff}}=\mathbf{K}_0+\mathbf{K}_{\mathrm{add}}$. The state-error identity and the linear residual relation give
\begin{equation}
\begin{aligned}
    \mathbf{e}^{n+1}
    &=
    \mathbf{P}\mathbf{e}^n
    +\mathbf{B}\boldsymbol{\delta}^{n+1},\\
    \boldsymbol{\delta}^{n+1}
    &=
    \mathbf{K}_{\mathrm{eff}}^{-1}
    \boldsymbol{\rho}^{n+1}.
\end{aligned}
\label{eq:supp-linear-error-recurrence}
\end{equation}
With zero initial error, repeated substitution yields
\begin{equation}
\begin{aligned}
    \mathbf{e}^n
    &=
    \mathbf{P}^{\,n}\mathbf{e}^0
    +
    \sum_{k=1}^{n}
    \mathbf{P}^{\,n-k}
    \mathbf{B}\boldsymbol{\delta}^k\\
    &=
    \sum_{k=1}^{n}
    \mathbf{P}^{\,n-k}
    \mathbf{B}\boldsymbol{\delta}^k\\
    &=
    \sum_{k=1}^{n}
    \mathbf{P}^{\,n-k}
    \mathbf{B}\mathbf{K}_{\mathrm{eff}}^{-1}
    \boldsymbol{\rho}^k.
\end{aligned}
\label{eq:supp-linear-error-representation}
\end{equation}
If $\sup_{j\geq0}\|\mathbf{P}^j\|_{\tau}\leq C_{\tau}$, then $\|\mathbf{e}^n\|_{\tau} \leq C_{\tau}b\sum_{k=1}^{n}d_{k-1}$. A sufficient condition for geometrically decaying powers is fixed-step average-acceleration Newmark with
\begin{equation}
    \mathbf{M}\succ0,
    \qquad
    \mathbf{K}_0\succ0,
    \qquad
    \mathbf{C}=\alpha_M\mathbf{M},
    \qquad
    \alpha_M>0.
\label{eq:supp-linear-decay-conditions}
\end{equation}
Under these conditions, all eigenvalues of $\mathbf{P}$ lie strictly inside the unit circle. Consequently, there exist constants $C_{\tau}\geq1$ and $0<r<1$ such that
\begin{equation}
    \|\mathbf{P}^{\,j}\|_{\tau}
    \leq C_{\tau}r^j,
    \qquad j\geq0.
\label{eq:supp-geometric-amplification-bound}
\end{equation}
If $d_{\max}:=\sup_{j\geq0}d_j<\infty$, Eq.~\eqref{eq:supp-linear-error-representation} then gives
\begin{equation}
\begin{aligned}
    \|\mathbf{e}^n\|_{\tau}
    &\leq
    C_{\tau}b
    \sum_{k=1}^{n}r^{n-k}d_{k-1}\\
    &\leq
    C_{\tau}b\,d_{\max}
    \frac{1-r^n}{1-r}\\
    &\leq
    \frac{C_{\tau}b\,d_{\max}}{1-r}.
\end{aligned}
\label{eq:supp-uniform-linear-error-bound}
\end{equation}
Under these conditions, error accumulation is uniformly bounded by $C_\tau b\,d_{\max}/(1-r)$, independently of the number of steps at a fixed $\Delta t$. The constants may depend on the time increment. In particular, choosing $\tau=\Delta t$ for fixed-step average-acceleration Newmark ($\beta=1/4$, $\gamma=1/2$) gives $b=\sqrt{21}$, but does not remove time-increment dependence from propagation or norm conversion.

To illustrate persistent bias, suppose that the fixed-step linear error dynamics are asymptotically stable, $\mathbf{K}_0\succ0$, and $\mathcal{G}_{\theta}(\mathbf{z};\mathbf{X})=\mathcal{S}(\mathbf{z})+\mathbf{b}$. Then $\boldsymbol{\delta}^n=\mathbf{b}$ and $\boldsymbol{\rho}^n=\mathbf{K}_{\mathrm{eff}}\mathbf{b}$, giving
\begin{equation}
    \mathbf{e}_u^{\infty}
    =
    \mathbf{K}_0^{-1}\mathbf{K}_{\mathrm{eff}}\mathbf{b}.
\label{eq:supp-persistent-bias-error}
\end{equation}
With mass-proportional damping, the modal amplitudes satisfy
\begin{equation}
    e_{u,j}^{\infty}=\left(1+\frac{a_0+a_1\alpha_M}{\omega_j^2}\right)b_j,
\label{eq:supp-modal-bias-amplification}
\end{equation}
where $\omega_j>0$ is the modal angular frequency. For average-acceleration Newmark, the leading amplification factor as $\Delta t$ decreases is $4/(\omega_j\Delta t)^2$. This occurs because a fixed displacement bias produces a residual load that grows with the effective stiffness. For a fixed residual, the local defect can instead decrease as $O(\Delta t^2)$ when the added mass term dominates. Error accumulation therefore depends on the magnitude, modal content, and temporal structure of the residual sequence, together with its propagation through the dynamics.

\subsection{Rollout time complexity}
\label{subsec:supp-complexity}

We consider the online arithmetic cost of \modelname{} after pretraining. Let $N$ denote the number of spatial points, proportional to the number of free degrees of freedom for a fixed number of components per node, and $T$ the number of time steps. Each step forms the effective inputs, evaluates the learned solver once, and recovers the dynamic state. With lumped mass and mass-proportional damping, forming the inputs and recovering the state require $\mathcal{O}(N)$ operations. The rollout cost is therefore
\begin{equation}
    C_{\mathrm{roll}}(N,T)
    =
    \mathcal{O}\!\left(
        T\,[C_{\mathrm{net}}(N)+N]
    \right),
\label{eq:supp-rollout-complexity}
\end{equation}
where $C_{\mathrm{net}}(N)$ is the cost of one network evaluation. The added-stiffness field is computed once for a fixed increment; prescribed variable increments require an $\mathcal{O}(N)$ update per step. Optional residual monitoring adds one restoring-force evaluation per monitored step.

Table~\ref{tab:supp-backbone-complexity} gives the operation counts for a standard FFT-based FNO on a uniform grid and a Transolver with $M\leq N$ slices, using $L$ layers and width $c$. The rollout scalings hold for fixed network depth, width, Fourier truncation, and slice count. They describe fixed architectures and do not establish scaling at a fixed error tolerance.

\begin{table}[h!]
    \centering
    \caption{Backbone evaluation and rollout complexity.}
    \label{tab:supp-backbone-complexity}
    {\renewcommand{\arraystretch}{1.15}%
    \begin{tabular}{@{}lll@{}}
        \toprule
        Backbone
        & One evaluation $C_{\mathrm{net}}(N)$
        & Rollout $C_{\mathrm{roll}}(N,T)$ \\
        \midrule
        FNO~\cite{fno}
        & $\mathcal{O}\!\left(L[cN\log N+c^2N]\right)$
        & $\mathcal{O}(TN\log N)$ \\
        \addlinespace
        Transolver~\cite{transolver}
        & $\mathcal{O}\!\left(L[cNM+cM^2+c^2N]\right)$
        & $\mathcal{O}(TN)$ \\
        \bottomrule
    \end{tabular}}
\end{table}

For comparison, a linear implicit FEM solver factorizes the effective stiffness at cost $F(N)$ and performs the subsequent triangular solves at cost $S(N)$. When the effective stiffness remains fixed, the factorization can be reused, giving
\begin{equation}
    C_{\mathrm{linear}}(N,T)
    =
    \mathcal{O}\!\left(
        F(N)+T[S(N)+N]
    \right).
\label{eq:supp-linear-fem-complexity}
\end{equation}
For standard 2D meshes with nested-dissection ordering, $F(N)=\mathcal{O}(N^{3/2})$ and $S(N)=\mathcal{O}(N\log N)$~\cite{generalized-nested,FEM-nested}. For 3D meshes, $F(N)=\mathcal{O}(N^{2})$ and $S(N)=\mathcal{O}(N^{4/3})$~\cite{generalized-nested}. In 2D, the cached per-step solve therefore has the same $\mathcal{O}(N\log N)$ order as FNO evaluation; this comparison alone does not determine their relative runtime. Changes in $\Delta t$ or the mass-proportional damping coefficient alter the effective stiffness and generally require a new numerical factorization unless its factors have already been cached. In \modelname{}, these changes require only an $\mathcal{O}(N)$ update of the added-stiffness field alongside the usual effective-input calculation, provided that the resulting inputs remain within the learned solver's applicable regime.

For nonlinear FEM using full Newton with sparse direct solves, the rollout cost is
\begin{equation}
    C_{\mathrm{Newton}}(N,T)
    =
    \mathcal{O}\!\left(
        T\left[
            N+\bar{k}_N
            \bigl(
                C_{\mathrm{asm}}(N)+F(N)+S(N)
            \bigr)
        \right]
    \right),
\label{eq:supp-newton-fem-complexity}
\end{equation}
where $\bar{k}_N$ is the average number of Newton iterations per step and $C_{\mathrm{asm}}(N)$ includes residual and tangent assembly. The numerical factors generally change with the Newton iterate, although the sparsity ordering can be reused. For fixed-order elements on the same 2D meshes, $C_{\mathrm{asm}}(N)=\mathcal{O}(N)$, giving $\mathcal{O}(T\bar{k}_N N^{3/2})$ work when at least one Newton iteration is performed per step. This compares with $\mathcal{O}(TN\log N)$ for FNO-based rollout and $\mathcal{O}(TN)$ for Transolver-based rollout under the fixed-architecture assumptions above.

\modelname{} replaces the iterative equilibrium solution with one network evaluation per step. Constitutive nonlinearity therefore introduces no additional online Newton iterations or tangent assemblies for a fixed architecture. Accuracy remains subject to the approximation, coverage, and propagation conditions in Supplementary Section~\ref{subsec:supp-error-decomp}.

These estimates exclude offline data generation and training. Measured runtimes and timing conventions are discussed in Section~\ref{subsubsec:classical-solvers}. 

\subsection{Differentiability of the rollout map}
\label{subsec:supp-differentiability}

For a prescribed rollout of $T$ steps, with geometry and pretrained weights fixed, write
\begin{equation}
\begin{aligned}
    \widehat{\mathbf{u}}^{n+1}
    &=
    \mathcal{G}_{\theta}\!\left(
        \mathcal{F}_{\mathrm{NM}}
        \bigl(
            \widehat{\mathbf{s}}^n,
            \mathbf{f}_{\mathrm{ext}}^{n+1};
            \mathbf{p}_n
        \bigr);
        \mathbf{X}
    \right),\\
    \widehat{\mathbf{s}}^{n+1}
    &=
    \mathcal{U}_{\mathrm{NM}}
    \bigl(
        \widehat{\mathbf{s}}^n,
        \widehat{\mathbf{u}}^{n+1};
        \mathbf{p}_n
    \bigr)
    =:
    \Phi_{\theta}
    \bigl(
        \widehat{\mathbf{s}}^n;
        \mathbf{f}_{\mathrm{ext}}^{n+1},
        \mathbf{p}_n
    \bigr),
\end{aligned}
\label{eq:supp-differentiable-step}
\end{equation}
where $\mathcal{F}_{\mathrm{NM}}$ forms the effective inputs and
$\mathbf{p}_n=(\mathbf{m},\alpha_M,\Delta t_n,\beta,\gamma)$.
The construction and recovery maps are smooth for $\Delta t_n>0$ and $\beta>0$. The rollout is therefore differentiable with respect to its initial state, force history, and exposed parameters wherever $\mathcal{G}_{\theta}$ is differentiable.

For a trajectory loss
$\mathcal{L}(\widehat{\mathbf{s}}^1,\ldots,\widehat{\mathbf{s}}^T)$,
retaining these operations in the computational graph permits reverse-mode automatic differentiation. With gradients written as column vectors, the adjoints
$\boldsymbol{\lambda}^n
=\mathrm{d}\mathcal{L}/\mathrm{d}\widehat{\mathbf{s}}^n$
satisfy
\begin{equation}
\begin{aligned}
    \boldsymbol{\lambda}^{T}
    &=
    \frac{\partial\mathcal{L}}
         {\partial\widehat{\mathbf{s}}^{T}},\\
    \boldsymbol{\lambda}^{n}
    &=
    \frac{\partial\mathcal{L}}
         {\partial\widehat{\mathbf{s}}^{n}}
    +
    \left(
        \frac{\partial\Phi_{\theta}}
             {\partial\widehat{\mathbf{s}}^{n}}
    \right)^{\mathsf T}
    \boldsymbol{\lambda}^{n+1},
    \qquad 1\leq n<T.
\end{aligned}
\label{eq:supp-rollout-adjoint}
\end{equation}
Here, the partial loss derivatives hold the other trajectory arguments fixed, and the step Jacobians are evaluated along the rollout. For example,
\begin{equation}
    \frac{\mathrm{d}\mathcal{L}}
         {\mathrm{d}\mathbf{f}_{\mathrm{ext}}^{n+1}}
    =
    \left(
        \frac{\partial\Phi_{\theta}}
             {\partial\mathbf{f}_{\mathrm{ext}}^{n+1}}
    \right)^{\mathsf T}
    \boldsymbol{\lambda}^{n+1},
    \qquad 0\leq n<T.
\label{eq:supp-force-gradient}
\end{equation}
Differentiation through these explicit operations requires no global linear solve, whereas implicit differentiation of an equilibrium equation involves a solve with the transposed effective tangent.

These are sensitivities of the surrogate rollout. The value-based bounds in Supplementary Note~\ref{subsec:supp-error-decomp} do not guarantee accurate Jacobians or reference sensitivities~\cite{derivative-no}. In particular, pretraining at a single $\kappa$, as for the nonlinear benchmarks, does not establish accuracy of derivatives in the added-stiffness direction. Here, differentiability characterizes the computational structure of \modelname{}; gradient accuracy and inverse identification are not evaluated.
\section{Problem specifications, data and reference solutions}
\label{sec:supp-data-protocols}

\subsection{Problem specifications}
\label{subsec:supp-problem-configuration}

Tables~\ref{tab:supp-benchmark-configurations} and~\ref{tab:supp-nonlinear-benchmark-configurations} summarize the physical parameters, static datasets, dynamic rollout settings, and neural architectures of all seven elastodynamic benchmarks.

\begin{table}[h!]
    \centering
    \caption{Configurations of the linear elastodynamic benchmarks. Operator inputs are the
    force and added-stiffness fields (three components each for the 3D solid).
    $N_t$ includes the initial state. GRF, Gaussian random field; DOFs, degrees of freedom;
    NR, not reported.}
    \label{tab:supp-benchmark-configurations}
    \ifdim\linewidth<145mm\scriptsize\else\footnotesize\fi
    \setlength{\tabcolsep}{2.5pt}
    \begin{tabularx}{\linewidth}{@{}>{\raggedright\arraybackslash}p{0.29\linewidth}*{5}{>{\centering\arraybackslash}X}@{}}
        \toprule
        & \begin{tabular}[b]{@{}c@{}}\textbf{1D}\\\textbf{truss}\end{tabular}
        & \begin{tabular}[b]{@{}c@{}}\textbf{2D}\\\textbf{panel}\end{tabular}
        & \begin{tabular}[b]{@{}c@{}}\textbf{2D}\\\textbf{perforated}\\\textbf{panel}\end{tabular}
        & \begin{tabular}[b]{@{}c@{}}\textbf{2D}\\\textbf{shear wall}\end{tabular}
        & \begin{tabular}[b]{@{}c@{}}\textbf{3D}\\\textbf{solid}\end{tabular} \\
        \midrule
        \multicolumn{6}{@{}l}{\textbf{\textit{Structure}}} \\
        \midrule
        \textbf{Size} [m]
        & 24 & $1\times1$ & $1\times1$ & $24\times36$ & 4 \\
        \textbf{Section}
        & {\footnotesize{$A=0.006$\,m$^2$}} & $t=0.01$\,m & $t=0.01$\,m
        & $t=0.25$\,m & {\footnotesize{$0.25\times0.25$\,m}} \\
        \textbf{Geometry}
        & -- & --
        & hole 
        & 12 floors & -- \\
        \textbf{Elas. modulus} $E$ [GPa]
        & 200 & 210 & 210 & 30 & 200 \\
        \textbf{Poisson's ratio} $\nu$
        & -- & 0.30 & 0.30 & 0.20 & 0.30 \\
        \textbf{Density} $\rho$ [kg\,m$^{-3}$]
        & 7,850 & 7,800 & 7,800 & 2,500 & 7,850 \\
        \textbf{DOFs}
        & 100 & 20,402 & 2,722 & 30,094 & 24,000 \\
        \midrule
        \multicolumn{6}{@{}l}{\textbf{\textit{Static pretraining}}} \\ \midrule
        \textbf{Samples}
        & 5,000 & 4,000 & 4,000 & 4,000 & 12,000 \\
        \textbf{Load amplitude} [kN]
        & \begin{tabular}[c]{@{}c@{}}mean 700\\s.d.\ 150\end{tabular}
        & \begin{tabular}[c]{@{}c@{}}GRF\\s.d.\ 33.33\end{tabular}
        & \begin{tabular}[c]{@{}c@{}}GRF\\s.d.\ 13.33\end{tabular}
        & \begin{tabular}[c]{@{}c@{}}scaled Beta\\$(1.2,3.5)$\end{tabular}
        & \begin{tabular}[c]{@{}c@{}}GRF mixed\\s.d.\ 0.0958\end{tabular}\\
        \textbf{Load range} [kN]
        & $[5,\,1{,}250]$ & $[-100,\,100]$ & $[-40,\,40]$
        & $[-100,\,100]$ & $[-0.29,\, 0.29]$ \\
        \textbf{Added stiff.} [MN\,m$^{-1}$]
        & $[10,2{\times}10^{4}]$
        & $[10^{-9},7{,}800]$
        & $[3.76,8.86]$
        & $[10^{4},4{\times}10^{4}]$
        & fixed field\\
        \midrule
        \multicolumn{6}{@{}l}{\textbf{\textit{Dynamic rollout}}} \\
        \midrule
        \textbf{Newmark} $\beta$
        & 0.25 & 0.25 & 0.25 & 0.25 & 0.2756 \\
        \textbf{Newmark} $\gamma$
        & 0.5 & 0.5 & 0.5 & 0.5 & 0.55 \\
        \textbf{Time step} $\Delta t$ [ms]
        & 1 & 20 & 10 & 10 & 10 \\
        \textbf{Duration} $T$ [s]
        & 1 & 7.98 & 9.99 & 1.99 & 2 \\
        \textbf{Time levels} $N_t$
        & 1,001 & 400 & 1,000 & 200 & 201 \\
        \textbf{Damping ratio} $\zeta$
        & 0.05 & 0.03 & 0.03 & 0.03 & $\zeta_1=0.03$ \\
        \midrule
        \multicolumn{6}{@{}l}{\textbf{\textit{Neural operator (\modelname{})}}} \\ \midrule 
        \textbf{Backbone}
        & Transolver & Transolver & Transolver & Transolver & FNO \\
        \textbf{Variant}
        & 1D & 2D & I & I & 3D \\
        \textbf{Layers}
        & 4 & 2 & 4 & 5 & 4 \\
        \textbf{Width}
        & 128 & 128 & 192 & 192 & 32 \\
        \textbf{Heads}
        & 4 & 8 & 8 & 8 & -- \\
        \textbf{Slices}
        & 16 & 32 & 32 & 32 & -- \\
        \textbf{MLP ratio}
        & 1 & 1 & 2 & 2 & -- \\
        \textbf{Fourier modes}
        & -- & -- & -- & -- & $(80,10,10)$ \\
        \bottomrule
    \end{tabularx}
\end{table}

\begin{table}[h!]
    \centering
    \caption{Configurations of the nonlinear (Neo-Hookean) elastodynamic benchmarks.
    $\mu$, shear modulus; $\bar{k}_{\mathrm{add}}$, added stiffness per unit reference length; $T_1$, first natural period;
    FC, fully connected layer; other entries as in Supplementary
    Table~\ref{tab:supp-benchmark-configurations}.}
    \label{tab:supp-nonlinear-benchmark-configurations}
    \ifdim\linewidth<145mm\scriptsize\else\footnotesize\fi
    \setlength{\tabcolsep}{2.5pt}
    \begin{tabularx}{\linewidth}{@{}>{\raggedright\arraybackslash}p{0.29\linewidth}*{2}{>{\centering\arraybackslash}X}@{}}
        \toprule
        & \begin{tabular}[b]{@{}c@{}}\textbf{1D}\\\textbf{Neo-Hookean bar}\end{tabular}
        & \begin{tabular}[b]{@{}c@{}}\textbf{2D}\\\textbf{Neo-Hookean panel}\end{tabular} \\
        \midrule
        \multicolumn{3}{@{}l}{\textbf{\textit{Structure}}} \\
        \midrule 
        \textbf{Size} [m]
        & 0.5 & $1\times1$ \\
        \textbf{Section}
        & $A=10^{-4}$\,m$^2$ & $t=0.01$\,m \\
        \textbf{Elastic constants}
        & $\mu=0.6$\,MPa & $E=1.5$\,MPa, $\nu=0.45$ \\
        \textbf{Density} $\rho$ [kg\,m$^{-3}$]
        & 1,030 & 1,100 \\
        \textbf{DOFs}
        & 101 & 2,178 \\
        \midrule
        \multicolumn{3}{@{}l}{\textbf{\textit{Static pretraining}}} \\
        \midrule 
        \textbf{Samples}
        & 7,600 & 6,000 \\
        \textbf{Static loads}
        & smooth spatial fields & spatial force fields \\
        \textbf{Added stiffness} 
        & $\bar{k}_{\mathrm{add}}=1.8$~MN\,m$^{-2}$ & fixed field \\
        \midrule
        \multicolumn{3}{@{}l}{\textbf{\textit{Dynamic rollout}}} \\
        \midrule 
        \textbf{Newmark} $\beta$
        & 0.25 & 0.25 \\
        \textbf{Newmark} $\gamma$
        & 0.5 & 0.5 \\
        \textbf{Time increment} $\Delta t$ [ms]
        & 0.48 ($T_1/100$) & 10 \\
        \textbf{Duration} $T$ [s]
        & $\approx0.1$ & 1 \\
        \textbf{Time levels} $N_t$
        & 201 & 101 \\
        \textbf{Damping ratio} $\zeta$
        & 0 & 0.02 \\
        \midrule
        \multicolumn{3}{@{}l}{\textbf{\textit{Neural operator (\modelname{})}}} \\
        \midrule 
        \textbf{Backbone}
        & FNO & FNO \\
        \textbf{Variant}
        & 1D & 2D \\
        \textbf{Layers}
        & 4 & 4 \\
        \textbf{Width}
        & 64 & 48 \\
        \textbf{Fourier modes}
        & 8 & $(12,12)$ \\
        \textbf{Padding}
        & 8 & 8 \\
        \textbf{FC width}
        & 171 & 128 \\
        \bottomrule
    \end{tabularx}
\end{table}

\subsubsection{Linear elastodynamics}
\label{subsubsec:supp-linear}

We consider five linear elastodynamic benchmarks spanning 1D, 2D, and 3D domains. Table~\ref{tab:supp-benchmark-configurations} summarizes their physical parameters, static datasets, dynamic rollout settings, and neural architectures. The 1D case uses uniaxial-stress elasticity, the 2D cases use plane-stress elasticity, and the 3D case uses 3D solid elasticity. In each benchmark, \modelname{} uses a static-pretrained operator to approximate the implicit displacement solve within the prescribed Newmark scheme.

\bmhead{1D linear}
The reference model uses two-node axial elements, with both ends fixed.  Dynamic tests include free release from a prescribed initial displacement, sustained distributed forcing, and forcing followed by release. For the configuration in Table~\ref{tab:supp-benchmark-configurations}, $\omega_1\Delta t\approx0.661$, where $\omega_1$ is the fundamental angular frequency of the undamped FE model with the prescribed constraints.

\bmhead{2D panel}
The reference model uses bilinear quadrilateral (Q4) elements, with both displacement components fixed along the bottom edge. A distributed vector load is prescribed through a smooth spatial profile and sinusoidal time dependence.

\bmhead{2D perforated panel}
The reference geometry contains a central circular opening and is discretized using three-node triangular (T3) elements. The bottom edge is fixed, and the domain is subjected to smooth distributed forcing.

\bmhead{2D shear wall}
The multistory perforated reference model employs T3 elements with a fixed bottom edge. Ramped two-frequency loading is applied in both coordinate directions. For this case and the perforated panel, Transolver-I denotes the Transolver on irregular meshes.

\bmhead{3D solid}
The reference model adopts fully integrated eight-node hexahedral (Hex8) elements on a $10\times10\times80$ nodal grid. All three displacement components are fixed on the $z=0$ face. Lumped mass and mass-proportional damping are used, with a damping ratio $\zeta_1=0.03$ at the fundamental frequency. The rollout starts from rest under distributed loading in all three coordinate directions. The load is ramped over $0.15\,\mathrm{s}$ and scaled to a peak of $13.675\,\mathrm{N}$ per node per component. This benchmark is evaluated at a fixed $\Delta t=0.01\,\mathrm{s}$ and $(\beta,\gamma)=(0.2756,0.55)$, corresponding to a fixed $\mathbf{K}_{\mathrm{add}}^{\mathrm{dyn}}$ and hence a fixed $\mathbf{K}_{\mathrm{eff}}=\mathbf{K}_0+\mathbf{K}_{\mathrm{add}}^{\mathrm{dyn}}$. In this configuration, $\omega_1\Delta t\approx0.808$.

\subsubsection{Nonlinear elastodynamics}
\label{subsubsec:supp-nonlinear}

The nonlinear benchmarks follow the formulation in Supplementary Note~\ref{subsubsec:supp-nonlinear-elastic}, with configurations summarized in Table~\ref{tab:supp-nonlinear-benchmark-configurations}.

\bmhead{1D nonlinear}
The case adopts the incompressible uniaxial Neo-Hookean reduction, with the left end fixed and the right end free. A one-cycle load burst is followed by free vibration. The added spring is specified through the stiffness density $\bar{k}_{\mathrm{add}}$ per unit reference length.

\bmhead{2D nonlinear}
The compressible Neo-Hookean example is modeled under plane strain, with both displacement components fixed along the bottom edge. Zero-force anchoring is applied to the neural displacement prediction.

\subsection{Spatial discretization convergence}
\label{subsec:supp-mesh-convergence}

\begin{figure}[h!]
    \centering
    \includegraphics[width=\linewidth]
    {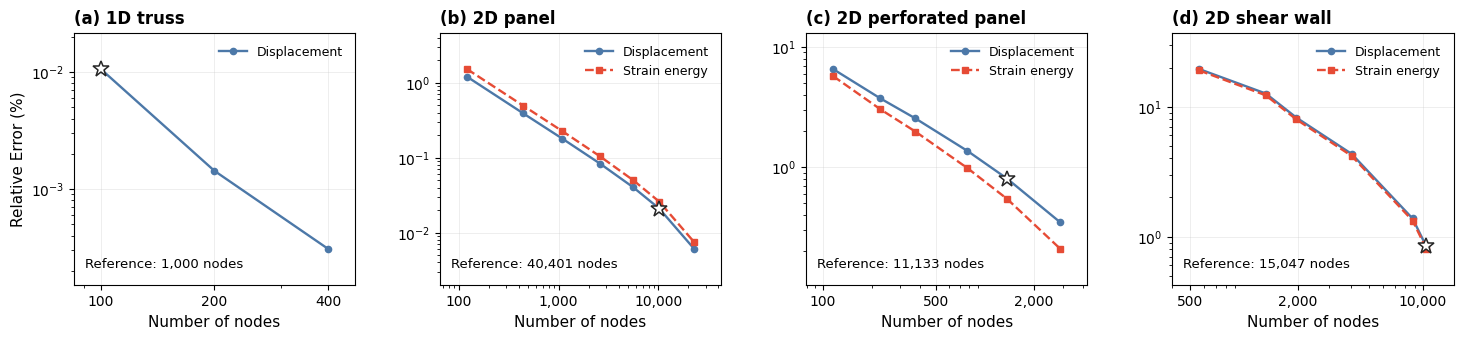}
    \caption{\textbf{Spatial discretization checks for linear benchmarks.}
    Relative differences from the finest reference mesh in each case:
    (\textbf{a}) displacement trajectories at a fixed time step;
    (\textbf{b-d}) displacement and strain energy from auxiliary
    linear-static tests. Stars highlight selected mesh resolutions.}
    \label{fig:supp-mesh-convergence}
\end{figure}

Figure~\ref{fig:supp-mesh-convergence} examines spatial refinement for the four linear cases. The 1D convergence check compares a representative displacement trajectory at a fixed time increment, while the 2D checks compare displacement and strain-energy quantities from auxiliary linear static tests. Errors are reported relative to the finest mesh considered. For the shear wall, the comparison between 10,478 and 15,047 nodes quantifies the change under refinement without estimating the remaining discretization error. Rollout errors are evaluated against the FE solution on the same discretization. Mesh convergence was also analyzed for the 3D solid. 

\subsection{Static pretraining data}
\label{subsec:supp-static-data}

Static pretraining datasets are generated as independent equilibrium problems for the benchmark systems specified in the previous supplementary section~\ref{subsec:supp-problem-configuration}. Each sample consists of a force field and an added spring stiffness input, with the corresponding equilibrium displacement as the target. These datasets contain no dynamic trajectories, velocities, or accelerations.

For the 2D examples and benchmarks, the nodal force fields in the $x$- and $y$-directions are sampled from spatially correlated Gaussian random fields. For node coordinates $\mathbf{x}_i$, the covariance is
\begin{equation}
    C_{ij}
    =
    \sigma_f^2
    \exp\left(
        -\frac{\|\mathbf{x}_i-\mathbf{x}_j\|}{\ell_c}
    \right)
\end{equation}
where $\sigma_f$ controls the force scale and $\ell_c$ is the spatial correlation length. The sampled fields are clipped to the prescribed static force range. Each input contains the two force components and a spatially repeated scalar added-stiffness channel, and the target contains the two displacement components. Representative static equilibrium inputs are shown in Fig.~\ref{fig:static_dynamic_coverage}a,d. Static dataset sizes and benchmark parameters are summarized in Tables~\ref{tab:supp-benchmark-configurations} and~\ref{tab:supp-nonlinear-benchmark-configurations}. For the 3D solid case, each input contains three force channels and three added-stiffness channels, with the three displacement components as the target prediction.

\subsection{Dynamic regimes}
\label{subsec:supp-dynamic-regimes}

The linear and nonlinear elastodynamic rollouts reported in the Results are evaluated zero-shot using \modelname{}, without training on dynamic trajectories. Loading conditions and time increments $\Delta t$ are selected to keep the resulting effective-force and added-stiffness inputs within the static-pretraining ranges.

For the neural-solver comparison, dynamic trajectory datasets are generated using the finite-element Newmark-$\beta$ solver. The IID training and test sets follow the same loading distribution and rollout horizon. OOD(F) uses external-force amplitudes outside the IID range at the same horizon, whereas OOD(H) extends the horizon while retaining the IID force range. These regimes are defined relative to the dynamic training distribution of the baselines; \modelname{} is evaluated zero-shot in all three regimes.

\subsection{Metrics}
\label{subsec:supp-metrics}

For static equilibrium tests, we report the relative $\ell_2$ error
\begin{equation}
    \mathcal{E}_{\mathrm{sta}}
    =
    \frac{
        \|\widehat{\mathbf{u}}-\mathbf{u}\|_2
    }{
        \|\mathbf{u}\|_2
    },
\end{equation}
where the norm is evaluated over the free degrees of freedom. For dynamic tests, we report the rollout relative $\ell_2$ error
\begin{equation}
    \mathcal{E}_{\mathrm{roll}}(T)
    =
    \frac{
        \|\widehat{\mathbf{U}}^{1:N_t-1}
        -\mathbf{U}^{1:N_t-1}\|_2
    }{
        \|\mathbf{U}^{1:N_t-1}\|_2
    },
\end{equation}
where $N_t$ is the number of time levels in the rollout, including the prescribed initial state $t^0$, so that $T=(N_t-1)\Delta t$ for a fixed increment. The initial state is excluded from the error. The displacement sequences are stacked over the evaluated time increments and free degrees of freedom, with all displacement components included jointly for 2D and 3D cases. Dynamic errors are evaluated against the classical Newmark-$\beta$ reference at the same time increment. For the cumulative-error curves in Figs.~2--4 and~6, we report
\begin{equation}
    E_{\mathrm{cum}}(n)=\frac{\|\widehat{\mathbf{U}}^{1:n}-\mathbf{U}^{1:n}\|_2}{\|\mathbf{U}^{1:N_t-1}\|_2},\qquad n=1,\dots,N_t-1,
\end{equation}
which is normalized by the full-rollout norm, is nondecreasing in $n$, and ends at $E_{\mathrm{roll}}(T)$. A constant per-step error therefore yields growth proportional to $\sqrt{n}$.

\subsection{Computational environment}
\label{subsec:supp-computational-environment}

In the manuscript and supplementary material, computations were performed on two platforms. The first was a cloud instance (Google Colab) with an NVIDIA A100 GPU (80 GB), an Intel Xeon processor (2.20 GHz, 12 CPUs), and 167 GB of system memory. The second was a local workstation at The University of Hong Kong with an NVIDIA GeForce RTX 4090 GPU (24 GB) and an Intel Core i9-13900KF processor (24 cores, 32 threads)

\modelname{} models for linear elastodynamic results (Fig. \ref{fig:linear-elastodynamics}) were trained and evaluated on the A100, and for the nonlinear elastodynamic results (Fig. \ref{fig:nonlinear-elastodynamics}) on the workstation RTX 4090. All comparisons for neural solvers (Fig. \ref{fig:neural-comparison} and Tables \ref{tab:neural-1d}-\ref{tab:neural-2d}) were also conducted on the RTX 4090. Runtime comparisons with classical solvers (Fig. \ref{fig:computational-efficiency}) were conducted on the cloud instance using 12 threads. The supplementary experiments were performed on the RTX 4090 workstation. 
\section{Neural solvers for dynamical systems}
\label{sec:supp-neural-methods}

\subsection{Baselines for learning dynamics}
\label{subsec:supp-dynamic-baselines}

We compare three baseline families trained on dynamic simulation data with \modelname{}, which learns from static equilibrium data. The baseline families differ in how they represent temporal dependence. Direct neural operators learn mappings from force histories to response histories, with longer trajectories evaluated through successive windows. Recurrent models carry temporal information in their recurrent memory. Autoregressive neural operators use explicit input histories, which are updated with predicted responses during rollout. Attention-based architectures are not treated as a separate family, since attention can be used within these formulations. In \modelname{}, a prescribed time integrator advances the physical state. The same neural-operator backbone can serve different roles.

\subsubsection{Direct neural operators}

Neural operators learn mappings between function spaces and have been applied to dynamic response prediction \cite{femu-fno,no-nonlinear-structure,nonlinear-operator-dynamics}. In our comparison, this baseline maps an external-force history to a displacement history,
\begin{equation}
    \widehat{\mathbf{U}}^{1:N_t}
    =
    \mathcal{B}_{\psi}
    \left(
        \mathbf{F}_{\mathrm{ext}}^{1:N_t},
        \mathbf{X}
    \right),
\end{equation}
where $N_t$ is the number of time samples in a prediction window, and $\mathbf{X}$ contains coordinates and other conditioning inputs where supplied. We primarily use FNO \cite{fno} as the direct neural-operator baseline. The mapping is trained for the prescribed benchmark initial conditions, without a separate initial-state input. Displacements are predicted jointly within each window without recursive feedback. Longer trajectories are evaluated through successive windows, with local time coordinates reset and no response state transferred between windows. Transolver \cite{transolver} can also be used as a direct neural operator with temporal embeddings. However, its autoregressive formulation performed better in our experiments. We therefore primarily use Transolver as an autoregressive neural-operator baseline in this comparison.

\subsubsection{Recurrent models}

For the recurrent baseline family, we use LSTM \cite{lstm} and ConvLSTM \cite{convlstm} to predict displacements in the 1D and 2D benchmarks, respectively. These models represent temporal dependence through a recurrent state that is updated at each time step \cite{deep-lstm,r-lstm-nonlinear-structure}.
\begin{equation}
    \left(
        \widehat{\mathbf{u}}^{n+1},
        \mathbf{z}^{n+1}
    \right)
    =
    \mathcal{R}_{\psi}
    \left(
        \widehat{\mathbf{u}}^{n},
        \mathbf{z}^{n},
        \mathbf{f}_{\mathrm{ext}}^{n+1},
        \mathbf{X}
    \right),
\end{equation}
Here, $\mathbf{z}^{n}$ collects the hidden and cell states. LSTM uses vector-valued features, whereas ConvLSTM uses spatial convolutions. Rollouts are initialized using the prescribed initial displacement and zero recurrent memory. During training, the LSTM baseline uses teacher forcing: the reference displacement $\mathbf{u}^{n}$ replaces $\widehat{\mathbf{u}}^{n}$ in the input. The selected ConvLSTM uses predicted-displacement feedback during training. During inference, both models advance autoregressively, passing predicted displacements and updated recurrent states to the next step.

\subsubsection{Autoregressive neural operators}

As mentioned previously, neural operators can learn to capture spatial or spatio-temporal information. Unlike direct neural operators, this baseline trains models using an autoregressive loss function to guide sequential prediction and capture temporal dependence. It differs from recurrent models in that no hidden states are carried between time steps.
\begin{equation}
    \widehat{\mathbf{u}}^{n+1}
    =
    \mathcal{A}_{\psi}
    \left(
        \widehat{\mathbf{U}}^{n-r+1:n},
        \mathbf{F}_{\mathrm{ext}}^{n-p+2:n+1},
        \mathbf{X}
    \right),
\end{equation}

where $r$ is the number of preceding displacement fields and $p$ is the number of force fields, including the excitation at the next prediction time. Each prediction updates the displacement history for the next step. Training uses reference histories or finite-step pushforward training, depending on the configuration. Within this family, both FNO and Transolver can serve as backbones. In our experiments, FNO did not perform well during autoregressive inference, whereas Transolver showed improved performance with this training formulation. We therefore distinguish this family from direct neural operators that learn spatio-temporal mappings.

\subsubsection{Physics-encoded neural operator}

\modelname{} uses a neural operator $\mathcal{G}_{\theta}$, trained to learn implicit solve, only on static equilibrium data, which predicts displacement from effective equilibrium inputs. The encoded Newmark-$\beta$ scheme constructs these inputs and advances the physical state, enabling dynamic rollout without training on dynamic trajectories (Section \ref{sec:method}, Eq.~\eqref{eq:modelname_operator_composition}).

\subsection{Model configurations and training settings}
\label{subsec:supp-neural-configurations}

Table \ref{tab:supp-neural-configurations} reports the selected architectures, initial learning rates, and trainable parameter counts for the neural-comparison experiments. Separate models are trained for the 1D and 2D benchmarks. The \modelname{} configurations for the main benchmarks are reported in the previous supplementary section \ref{subsec:supp-problem-configuration}.

\begin{table}[htbp]
    \centering
    \caption{\textbf{Neural model configurations.}
    All models are trained using the Adam optimizer.}
    \label{tab:supp-neural-configurations}
    \small
    \setlength{\tabcolsep}{4pt}
    \renewcommand{\arraystretch}{1.0}
    \begin{tabular}{@{}p{0.30\linewidth}
                       p{0.48\linewidth}r@{}}
        \toprule
        Model & Configuration & Parameters \\
        \midrule

        \multicolumn{3}{@{}l}{\textit{1D dynamic benchmark}} \\
        \midrule[0.2pt]

        FNO-2D
        & \texttt{n\_layers}=4 \newline
          \texttt{hidden\_channels}=64 \newline
          \texttt{n\_modes}=$(16,16)$
        & 2,410,177 \\
        \midrule[0.2pt]

        Transolver-1D
        & \texttt{n\_layers}=4 \newline
          \texttt{n\_hidden}=192 \newline
          \texttt{n\_head}=8 \newline
          \texttt{slice\_num}=32 \newline
          \texttt{unroll\_steps}=4
        & 1,127,393 \\
        \midrule[0.2pt]

        LSTM
        & \texttt{num\_layers}=2 \newline
          \texttt{hidden\_size}=512
        & 3,877,476 \\
        \midrule[0.2pt]

        \textbf{\modelname{} (FNO-1D)}
        & \texttt{n\_layers}=4 \newline
          \texttt{hidden\_channels}=64 \newline
          \texttt{n\_modes}=$(16,)$
        & 198,337 \\
        \midrule[0.2pt]

        \textbf{\modelname{} (Transolver-1D)}
        & \texttt{n\_layers}=2 \newline
          \texttt{n\_hidden}=128 \newline
          \texttt{n\_head}=8 \newline
          \texttt{slice\_num}=8
        & 202,401 \\

        \midrule
        \multicolumn{3}{@{}l}{\textit{2D dynamic benchmark}} \\
        \midrule[0.2pt]

        FNO-3D
        & \texttt{n\_layers}=4 \newline
          \texttt{hidden\_channels}=32 \newline
          \texttt{n\_modes}=$(32,16,16)$
        & 18,887,842 \\
        \midrule[0.2pt]

        Transolver-2D
        & \texttt{n\_layers}=4 \newline
          \texttt{n\_hidden}=128 \newline
          \texttt{n\_head}=8 \newline
          \texttt{slice\_num}=32
        & 1,421,986 \\
        \midrule[0.2pt]

        ConvLSTM
        & \texttt{hidden}=$[64,64,64,64]$ \newline
          \texttt{kernel\_size}=3
        & 1,047,170 \\
        \midrule[0.2pt]

        \textbf{\modelname{} (FNO-2D)}
        & \texttt{n\_layers}=4 \newline
          \texttt{hidden\_channels}=32 \newline
          \texttt{n\_modes}=$(16,16)$
        & 603,234 \\
        \midrule[0.2pt]

        \textbf{\modelname{} (Transolver-2D)}
        & \texttt{n\_layers}=2 \newline
          \texttt{n\_hidden}=128 \newline
          \texttt{n\_head}=8 \newline
          \texttt{slice\_num}=16
        & 727,858 \\

        \bottomrule
    \end{tabular}
\end{table}

The LSTM uses dropout of $0.1$, and ConvLSTM uses residual prediction with layerwise dilations of $(1,2,4,8)$. The initial learning rate is $10^{-3}$, except for ConvLSTM, which uses $3\times10^{-4}$. Direct and recurrent baselines use batch sizes of 16 in 1D and 8 in 2D. Autoregressive Transolver models use a batch size of 128, and \modelname{} uses 32. All runs use random seed 42.

Training loss is monitored for convergence, and checkpoints are selected by validation relative $\ell_2$ error. Baselines use IID dynamic validation, with autoregressive rollout where applicable; \modelname{} uses static validation.

Direct trajectory models use windows of 50 time samples in 1D and 100 in 2D. Longer trajectories are processed through successive windows with local time coordinates reset and no state transferred between windows. Model weights and normalization statistics remain fixed during evaluation.

\subsection{Benchmark regimes and coverage}
\label{subsec:supp-data-coverage}

\begin{figure}[htbp]
    \centering
    \includegraphics[width=\linewidth]
    {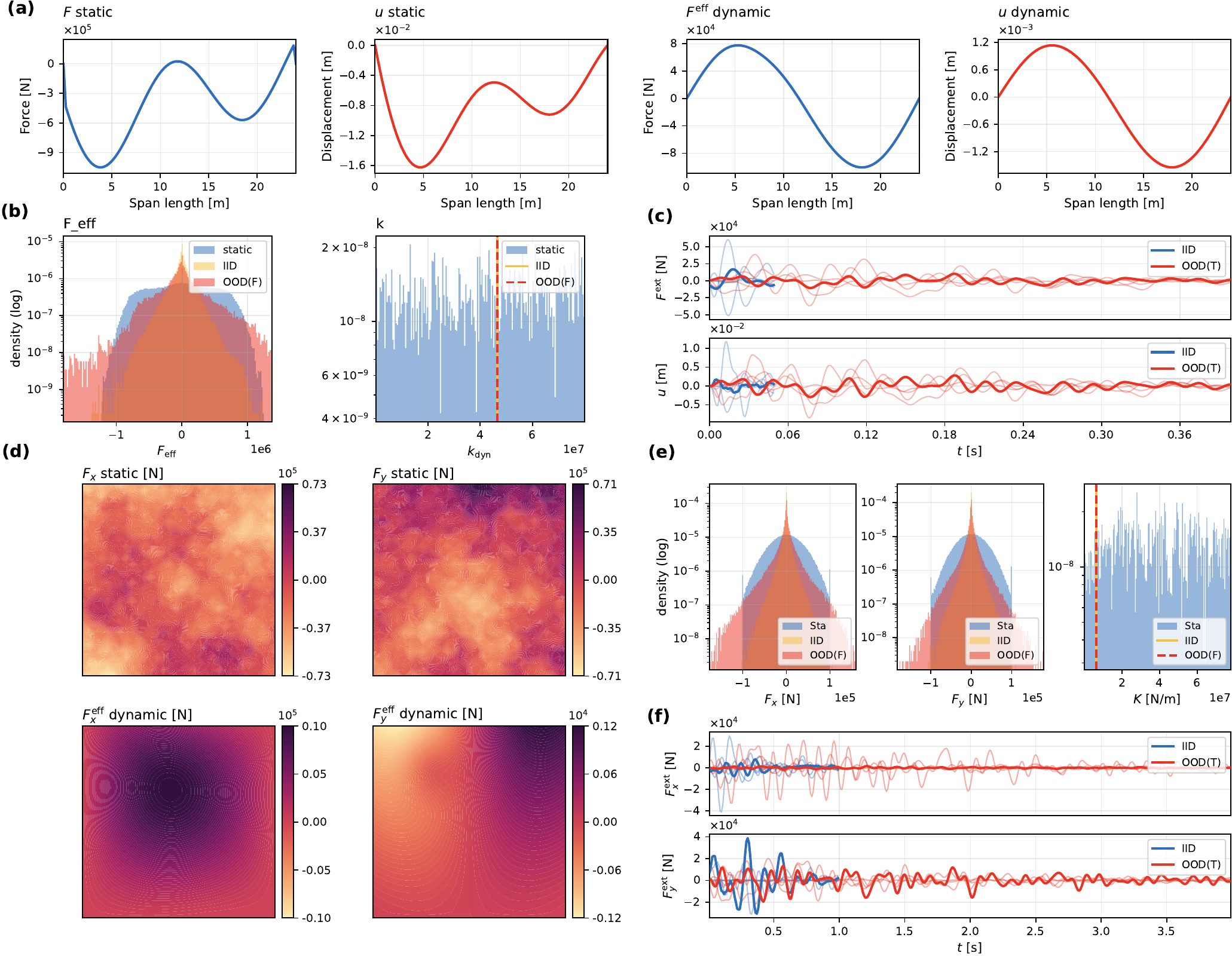}
    \caption{\textbf{Static-pretraining coverage and
    dynamic evaluation regimes.}
    (\textbf{a},\textbf{d}) Representative static and
    dynamic effective-equilibrium data for the 1D and
    2D benchmarks, respectively.
    (\textbf{b},\textbf{e}) Empirical force and stiffness
    input distributions for static pretraining,
    IID dynamics and OOD(F), with dynamic forces
    expressed as effective-equilibrium inputs.
    (\textbf{c},\textbf{f}) Representative time histories
    for IID and OOD(H). (labeled OOD(T) in the legends).}
    \label{fig:static_dynamic_coverage}
\end{figure}

Dynamic baselines are trained on IID trajectories and selected using IID validation data. Evaluation uses held-out IID trajectories and two OOD regimes: OOD(F) increases external-force amplitudes beyond the IID range at the training horizon, whereas OOD(H) extends the horizon while retaining the IID amplitude range (Fig.~\ref{fig:static_dynamic_coverage}).

For a controlled comparison, the static pretraining data for \modelname{} are designed to cover the effective-force and stiffness input ranges associated with IID dynamics. The model is evaluated zero-shot on the same IID and OOD trajectories, without dynamic fine-tuning. The strong generalization reported in Section~\ref{sec:results} motivates its investigation for other dynamical systems.

\section{Effective-stiffness coverage governs static-to-dynamic solver reuse}
\label{sec:supp-coverage}

\subsection{Effective-stiffness coverage condition}
\label{subsec:supp-coverage-condition}

This supplementary experiment examines whether a neural solver trained only on static augmented-equilibrium systems can be reused as the implicit displacement solver inside Newmark-$\beta$ dynamics. The static training problem follows the augmented equilibrium form $(K_0 + K_{\mathrm{spring}})u = F$, where the stiffness augmentation is represented in the model input as the scalar channel $[f_x, f_y, \mathrm{total\_k}]$ repeated over all nodes. This supplementary experiment has two objectives: (i) to evaluate variable-$\Delta t$ dynamic rollout using a single static-pretrained neural operator, and (ii) to test whether successful static-to-dynamic reuse is governed by coverage of the Newmark effective stiffness within the static stiffness support used during pretraining.

At each Newmark step, the implicit displacement update can be written as
\begin{equation}
    K_{\mathrm{eff}}(\Delta t) u^{n+1}
    =
    F_{\mathrm{eff}}^{n+1}
\end{equation}
where
\begin{equation}
    K_{\mathrm{eff}}(\Delta t)
    =
    K_0
    +
    \frac{1}{\beta \Delta t^2} M
    +
    \frac{\gamma}{\beta \Delta t} C 
\end{equation}

For the mass-proportional damping used in this study, $C=\alpha_M M$. The Newmark-induced stiffness contribution can therefore be mapped to the same scalar input form used in static pretraining: $\mathrm{total\_k}_{\mathrm{input}}(\Delta t)= N_{\mathrm{nodes}} m_{\mathrm{node}}\left(a_0 + a_1 \alpha_M\right)$. We define the input-level coverage ratio as
\begin{equation}
    \rho_{\mathrm{input}}(\Delta t)
    =
    \frac{
    \mathrm{total\_k}_{\mathrm{input}}(\Delta t)
    }{
    \mathrm{total\_k}_{\max,\mathrm{static}}
    }
\end{equation}

Thus, $\rho_{\mathrm{input}}<1$ indicates that the Newmark-induced effective stiffness lies within the static pretraining support, while $\rho_{\mathrm{input}}>1$ corresponds to extrapolation beyond the static stiffness range. This coverage ratio provides a direct diagnostic for when static-to-dynamic reuse is expected to remain reliable.

\subsection{Variable-$\Delta t$ experiment design}
\label{subsec:supp-coverage-experiment-design}

We evaluated the coverage condition in section \ref{subsec:supp-coverage-condition} on the $32\times32$ Panel2D benchmark with Q4 plane-stress finite elements. The bottom edge was fixed, and all other nodes were free. A single static-pretrained Transolver2D checkpoint was reused for all dynamic evaluations; no dynamic trajectory training or $\Delta t$-specific retraining was performed.

\begin{figure}[h!]
    \centering
    \includegraphics[width=\linewidth]{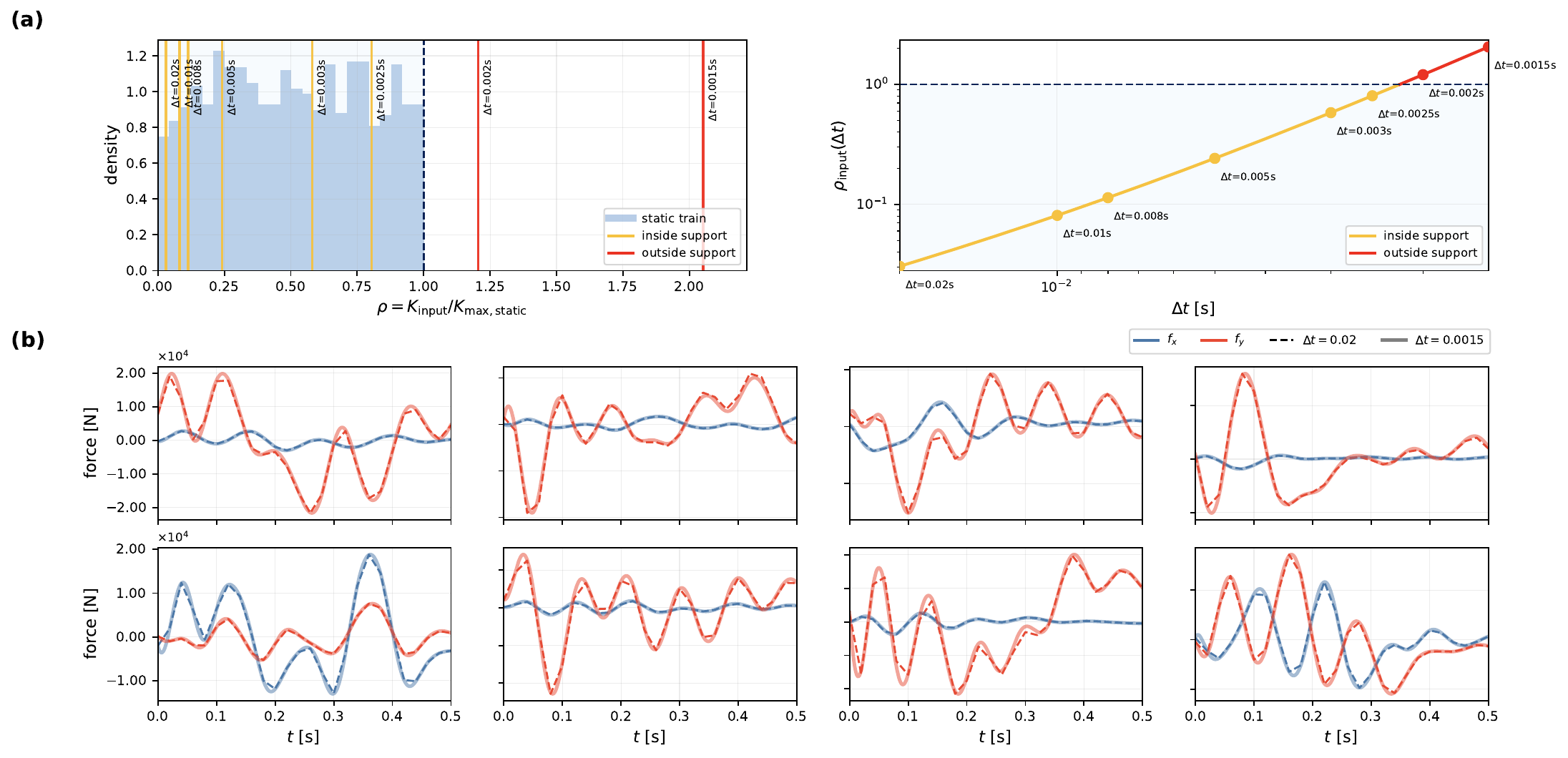}
    \caption{Coverage-controlled design of the static-to-dynamic transfer experiment.  \textbf{(a)} Distribution of static stiffness inputs and Newmark-induced effective-stiffness ratios across the tested time steps. \textbf{(b)} Dynamic cases (applied forces), shown through the nodal components $f_x$ and $f_y$ at a representative node on the $32\times32$ panel under $\Delta t=0.02$ s and $\Delta t=0.0015$ s.}
    \label{fig:supp_s3_design}
\end{figure}

As depicted in Figure~\ref{fig:supp_s3_design}, the target duration was fixed at $T=0.5\,\mathrm{s}$, while the Newmark time step was varied as $\Delta t=\{0.020,\ 0.010,\ 0.008,\ 0.005,\ 0.003,\ 0.0025,\ 0.002,\ 0.0015\}\,\mathrm{s}$. These values span an input-level coverage range from deeply inside the static support to clearly outside it, with $\rho_{\mathrm{input}}=0.031$--$2.051$. The first six cases satisfy $\rho_{\mathrm{input}}\leq0.804$, while the last two exceed the static support with $\rho_{\mathrm{input}}=1.205$ and $2.051$.

Synthesized spatial--temporal force histories were applied to the $32\times32$ Panel2D nodal grid for eight dynamic cases. Each case used separable-type nodal force fields $f_x(x,y,t) = A_x s_x(t)\phi_x(x,y), f_y(x,y,t) = A_y s_y(t)\phi_y(x,y)$ where $s_x(t)$ and $s_y(t)$ are continuous multi-sinusoidal temporal signals with pulse-like components, and $\phi_x(x,y)$ and $\phi_y(x,y)$ are mixed spatial load patterns. For each $\Delta t$, the same continuous force cases were resampled on the corresponding time grid, so that the physical loading event was held fixed while the Newmark discretization, number of implicit updates, and effective stiffness changed.

For each force case and time step, we computed the Newmark reference solution, the \modelname{} one-step solution using reference states, and the \modelname{} rollout solution using recursively predicted states. We report the full-duration error over the complete event and a fixed-prefix error over the first 25 Newmark updates. The former evaluates physical rollout accuracy, while the latter controls for the number of neural implicit-solver calls.

\subsection{Stable rollout within static support}
\label{subsec:supp-coverage-result}

\subsubsection{Stability and generalization} 
The static-pretrained \modelname{} solver remains stable when the Newmark-induced effective stiffness is within the static support. For $\rho_{\mathrm{input}}\leq0.804$, the full-duration rollout error remains nearly constant, $0.01397$--$0.01520$, while the number of Newmark updates increases from $25$ to $200$ (Table~\ref{tab:supp_s3_variable_dt}). The solver also remains accurate under extrapolation at $\Delta t=0.002$ s, where $\rho_{\mathrm{input}}=1.205$ and the full-rollout error is $0.01836$. These results indicate that the learned static solver can be repeatedly reused across coarse and fine time steps inside and beyond the covered stiffness range.

\subsubsection{Rollout robustness} 
\begin{figure}[h!]
    \centering
    \includegraphics[width=\linewidth]{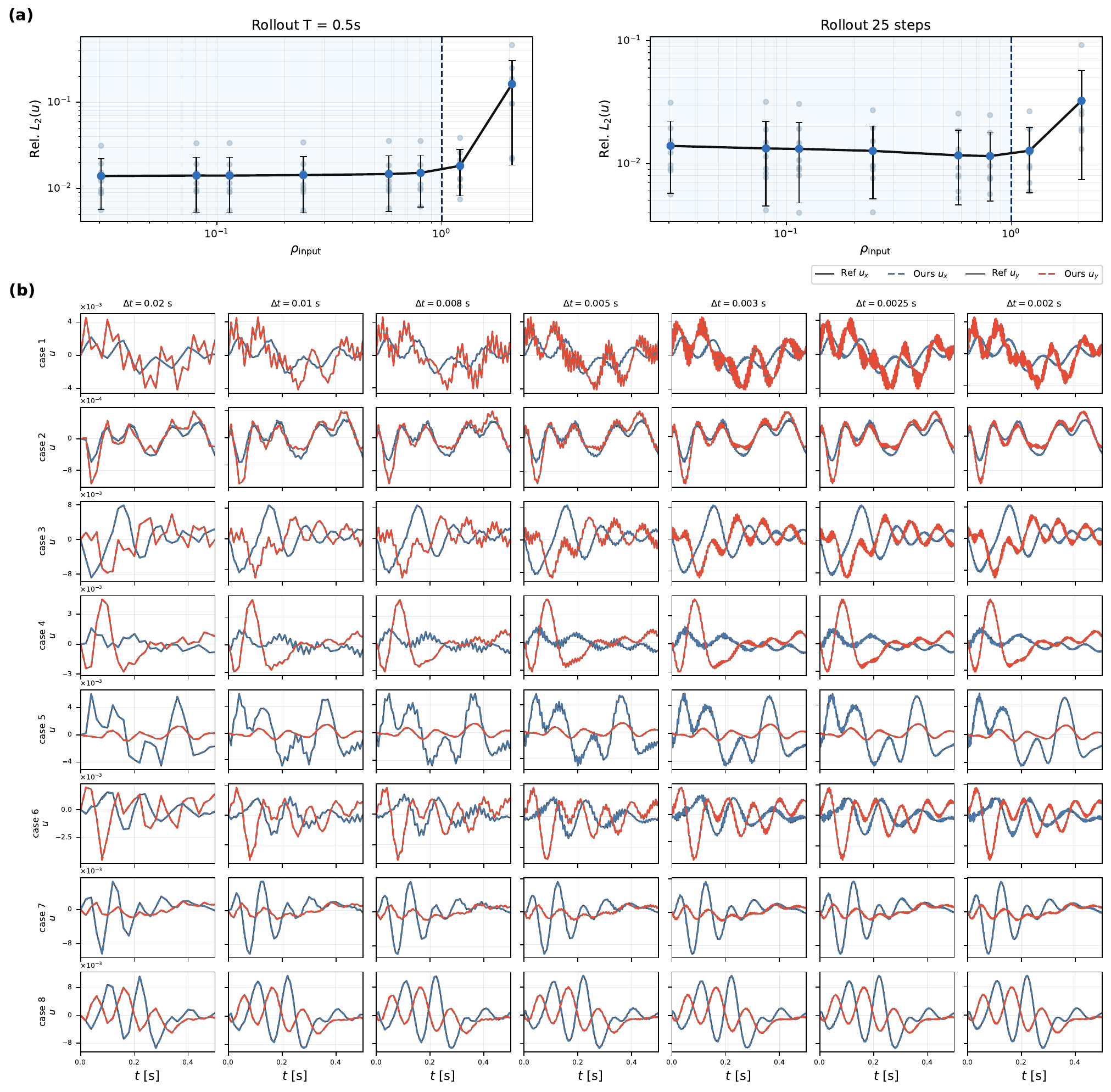}
    \caption{Variable-$\Delta t$ rollout accuracy of the static-pretrained \modelname{} solver.
    \textbf{(a)} rollout errors versus coverage ratio $\rho_{\mathrm{input}}$. \textbf{(b)} Representative displacement responses.}
    \label{fig:supp_s3_results}
\end{figure}

The fixed-depth diagnostic further isolates local solver reuse from trajectory length. Across the covered regime, the 25-step rollout error stays bounded between $0.01155$ and $0.01397$, showing that reducing $\Delta t$ does not degrade the local implicit solve as long as the effective stiffness remains within support. Meanwhile, the full $T\approx0.5$ s rollout at $\Delta t=0.002$ s remains accurate over $250$ recursive updates, supporting autoregressive stability over long integration horizons (Fig.~\ref{fig:supp_s3_results}).

\subsubsection{Coverage-driven failure mode} 
The dominant failure mode appears when the effective stiffness moves far outside the static dataset range. At $\Delta t=0.0015$ s, $\rho_{\mathrm{input}}=2.051$, and the full-rollout error increases to $0.16280$. This transition is consistent with the Newmark mass contribution scaling as $1/\Delta t^2$: sufficiently small time steps induce effective stiffnesses not seen during static pretraining. Thus, the limitation is not the Newmark rollout formulation itself, but insufficient static-support coverage for the induced dynamic solves.

\begin{table}[h!]
\centering
\caption{Variable-$\Delta t$ rollout accuracy of the static-pretrained \modelname{} solver.}
\label{tab:supp_s3_variable_dt}
\vspace{-0.4em}
\scriptsize
\setlength{\tabcolsep}{4pt}
\begin{tabular*}{\linewidth}{@{\extracolsep{\fill}}ccccc}
\toprule
$\Delta t$ [s] & $N$ & $\rho_{\mathrm{input}}$ & $e_{\mathrm{roll}}^{T}$ & $e_{\mathrm{roll}}^{25}$ \\
\midrule
0.0200 & 25  & 0.031 & 0.01397 & 0.01397 \\
0.0100 & 50  & 0.081 & 0.01416 & 0.01332 \\
0.0080 & 63  & 0.114 & 0.01414 & 0.01320 \\
0.0050 & 100 & 0.242 & 0.01433 & 0.01273 \\
0.0030 & 167 & 0.581 & 0.01473 & 0.01170 \\
0.0025 & 200 & 0.804 & 0.01520 & 0.01155 \\
0.0020 & 250 & 1.205 & 0.01836 & 0.01277 \\
0.0015 & 333 & 2.051 & 0.16280 & 0.03253 \\
\bottomrule
\end{tabular*}
\vspace{-0.8em}
\end{table}

\subsection{Million-step rollout}
\label{subsec:supp-million-step}

We use the 100-node 1D linear elastodynamic system in Table~\ref{tab:supp-benchmark-configurations}, initialized at rest, with mass-proportional damping specified by $\zeta_1=0.02$. The static-pretrained \modelname{} with a Transolver backbone performs $10^6$ consecutive Newmark updates at a fixed time increment $\Delta t=10^{-3}\,\mathrm{s}$, corresponding to $1{,}000\,\mathrm{s}$ of simulated time. The loading comprises 100 blocks of 10,000 updates, each containing six episodes of chirps, multisines, random multisines, or bursts with carrier frequencies of 3-18 Hz. Spatial patterns combine Gaussian patches and sinusoidal shapes. Each episode is followed by 60 updates with zero external force. Network weights remain fixed, and displacement, velocity, and acceleration are propagated continuously using the predicted states, without resets, teacher forcing, or reference-state corrections.

The reference trajectory is computed using a finite-element Newmark solver with LU factorization and identical mesh, loading, damping, and time increment. Relative $\ell_2$ displacement errors are evaluated over all nodes and updates within each evaluation interval. Errors over the first $10^4$, $10^5$ and $10^6$ updates are $0.586\%$, $0.581\%$ and $0.579\%$, respectively. Across the 100 non-overlapping windows of 10,000 updates, the median and maximum errors are $0.589\%$ and $0.685\%$. These results provide empirical evidence of sustained rollout accuracy relative to the discrete Newmark reference under the tested conditions.
\section{Spatial resolution transfer}
\label{sec:supp-resolution-transfer}

We evaluate spatial resolution transfer in the 1D linear elastodynamic system using pretrained \modelname{} operators at the native, $2\times$ and $4\times$ resolutions without retraining. Compared with the finite-element reference at each resolution, the rollout relative $\ell_2$ errors are $0.29\%$, $7.96\%$ and $13.00\%$ for the FNO backbone, and $0.21\%$, $12.13\%$ and $18.65\%$ for Transolver, respectively. Both operators can be evaluated on finer meshes, but accuracy decreases under the tested transfer. Super-resolution remains a subject for further investigation.

\begin{figure}[h!]
    \centering
    \includegraphics[width=\linewidth]
    {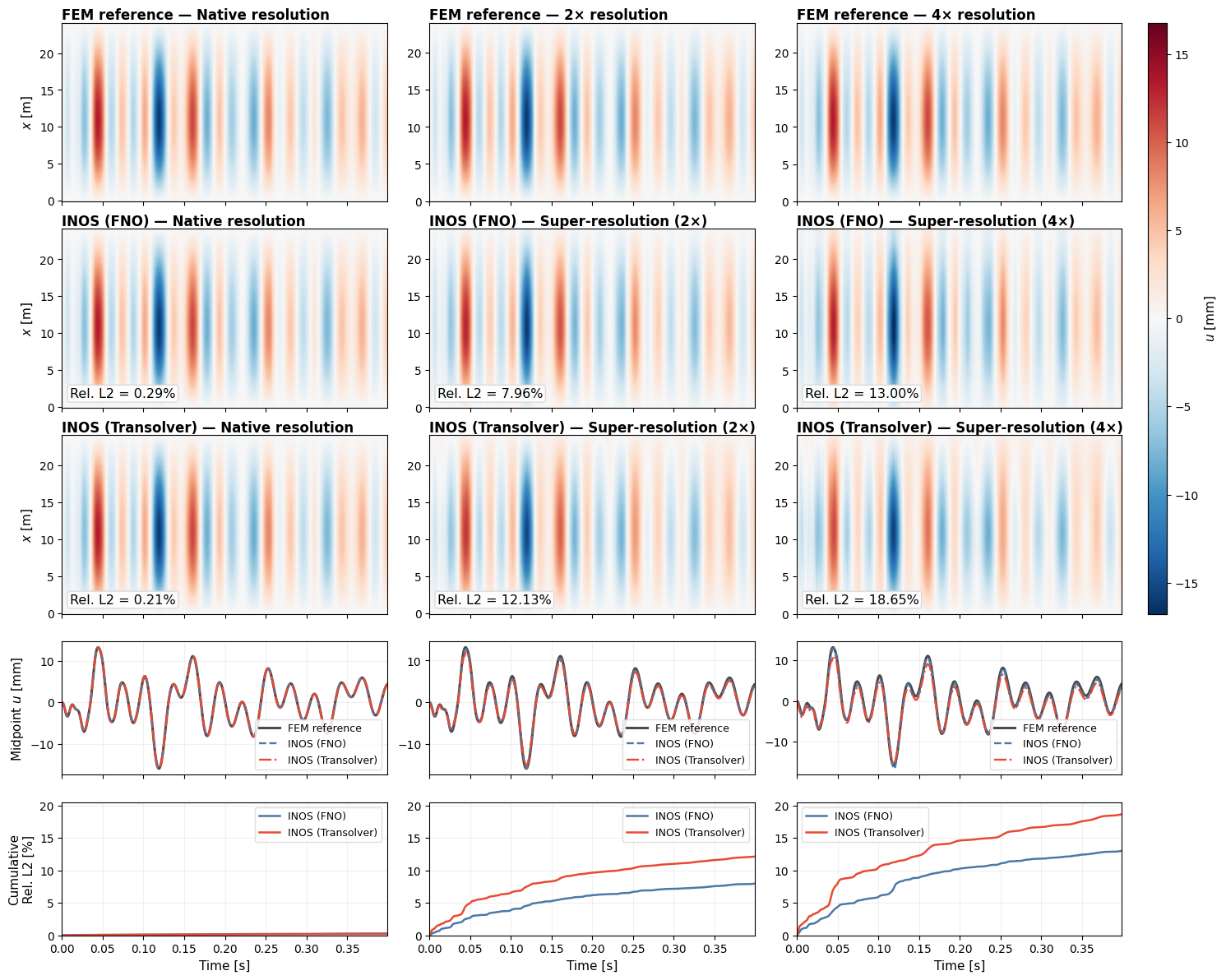}
    \caption{\textbf{Spatial resolution transfer in 1D linear elastodynamics.} Columns show the native, $2\times$ and $4\times$ resolutions. Rows 1-3 show displacement fields from the FE reference and \modelname{} with FNO and Transolver backbones. Rows 4-5 show midpoint displacement histories and cumulative errors, respectively. Annotations report rollout relative $\ell_2$ errors.}
    \label{fig:supp-resolution-transfer}
\end{figure}

\end{appendices}


\end{document}